\documentclass[runningheads]{llncs}
\usepackage[T1]{fontenc}
\usepackage{graphicx}
\usepackage{float}
\usepackage{enumerate}
\usepackage{xcolor}
\usepackage{algorithm}
\usepackage{algorithmic}
\usepackage{amsmath}
\usepackage{amsfonts}

\usepackage{amssymb}
\usepackage{hyperref}
\usepackage{multirow}
\usepackage{listings}
\usepackage{eurosym}
\usepackage{lstautogobble}
\usepackage{url}
\usepackage{tikz}
\usepackage{siunitx}
\usepackage{xspace}
\usepackage{pgfplots}
\usepackage{booktabs}
\usepackage{bm}
\usepackage{mathtools}
\usepackage{epstopdf}
\usepackage{lipsum}
\usepackage{placeins}
\usepackage{soul}
\usepackage{pifont}
\usepackage{etoolbox}
\usepackage{pgf}
\usepackage{array}
\usepackage[multiple]{footmisc}
\usepackage[most,skins,theorems]{tcolorbox}
\usepackage{colortbl}
\usepackage{collcell}
\usepackage{subcaption}
\usepackage{wrapfig}
\usepackage{tabularx}
\usepackage{makecell}
\usepackage{microtype}
\usepackage{fvextra}
\RecustomVerbatimCommand{\VerbatimInput}{VerbatimInput}{%
  fontsize=\footnotesize,
  breaklines=true,
  breakanywhere=true,
  obeytabs=true
}

\usetikzlibrary{arrows,backgrounds,positioning,automata}

\definecolor{dkgreen}{rgb}{0,0.6,0}
\definecolor{gray}{rgb}{0.5,0.5,0.5}
\definecolor{mauve}{rgb}{0.58,0,0.82}
\definecolor{red}{rgb}{1,0,0}
\definecolor{darkblue}{rgb}{0,0,0.5}

\newcommand{\cf}{{cf.~}}

\newcommand{\red}[1]{\textcolor{red}{#1}}

\newcolumntype{L}{>{\arraybackslash}m{.45\columnwidth}|}

\newcommand{\owner}{\protect {$\mathcal{O}$}\xspace}
\newcommand{\adversary}{\protect {$\mathcal{A}$}\xspace}

\graphicspath{ {./images/} }

\tcbuselibrary{listingsutf8}

\newtcolorbox{resultbox}[1][]{%
  colback=green!5!white,
  colframe=green!70!black,
  fonttitle=\bfseries,
  #1
}
\newtcolorbox{hypobox}[1][]{%
  colback=red!5!white,
  colframe=red!80!black,
  fonttitle=\bfseries,
  #1
}

\definecolor{LightGreen}{rgb}{0.88, 1.0, 0.88}
\definecolor{LightBlue}{rgb}{0.90, 0.95, 1.0}

\definecolor{tableblue}{RGB}{44,127,184}
\definecolor{mydeepblue}{RGB}{64, 82, 181}
\definecolor{myyellow}{RGB}{245, 236, 66}
\definecolor{mylightgreen}{RGB}{204,255,204}
\definecolor{mylightblue}{RGB}{204,229,255}
\newcommand{\cellpercent}[1]{%
  \pgfmathsetmacro\percent{int(max(0,min(100*(#1)/\maxval,100)))}%
  {%
    \expandafter\cellcolor\expandafter{%
      mylightgreen!\percent!mylightblue%
    }%
    $#1$%
  }%
}

\tcbset{
  aibox/.style={
    width=\linewidth,
    top=8pt,
    bottom=4pt,
    colback=blue!6!white,
    colframe=black,
    colbacktitle=black,
    enhanced,
    center,
    attach boxed title to top left={yshift=-0.1in,xshift=0.15in},
    boxed title style={boxrule=0pt,colframe=white,},
  },
  notebox/.style={
    colback=yellow!20!white,
    colframe=yellow!75!black,
    coltitle=black,
    enhanced,
    center,
    attach boxed title to top left={yshift=-0.1in,xshift=0.15in},
    boxed title style={
    colback=yellow!60!white,
    },
    title=Note,
  }
}
\newtcolorbox{AIbox}[2][]{aibox,title=#2,#1}
\newtcolorbox{notebox}[2][]{notebox,title=#2,#1}

\definecolor{high}{HTML}{76f013}
\definecolor{low}{HTML}{ec462e}
\newcommand*{\opacity}{90}
\newcommand*{\minval}{0.0}
\newcommand*{\maxval}{100.0}
\newcommand{\gradient}[1]{
    $\ifdimcomp{#1pt}{>}{\maxval pt}{#1}{
        \ifdimcomp{#1pt}{<}{\minval pt}{#1}{
            \pgfmathparse{int(round(100*(#1/(\maxval-\minval))-(\minval*(100/(\maxval-\minval)))))}
            \xdef\tempa{\pgfmathresult}
            \cellcolor{high!\tempa!low!\opacity} #1$
    }}
}

\pgfplotsset{compat=1.18}

\begin{document}
\title{Merge Now, Regret Later:\\ The Hidden Cost of Model Merging Is Adversarial Transferability}
\titlerunning{Merge Now, Regret Later}
\author{Mauro Conti\inst{1, 2} \and
Ankit Gangwal\inst{1, 3} \and
Aaryan Ajay Sharma\inst{3, 4,}$^\dagger$}
\authorrunning{M. Conti, A. Gangwal, A. A. Sharma}
\institute{
    University of Padua, Italy\\
    \and
    Örebro University, Sweden\\
    \and
    IIIT Hyderabad, India\\
    \and
    University of Twente, Netherlands\\
    \email{mauro.conti@unipd.it}, \email{gangwal@iiit.ac.in}, \email{aaryan.sharma@utwente.nl}
}
\maketitle
{\let\thefootnote\relax\footnotetext{Authors ordered alphabetically by last name.}}
{\let\thefootnote\relax\footnotetext{$^\dagger$Corresponding author.}}

\setcounter{footnote}{0}

\begin{abstract}
    {Model Merging~(MM) has proven to be an effective alternative to multi-task learning, where several fine-tuned models are merged, without access to {the} tasks' training data, into one model that retains performance across different tasks.} {Recent works have explored the security of MM, showing how MM can confer robustness against various adversarial attacks.} {However, none of them {has} sufficiently explored its impact on transfer attacks using transferable adversarial examples.} 
    
    In this work, we study the effect of MM on the transferability of adversarial examples. We perform comprehensive evaluations and statistical analysis consisting of eight MM methods, seven datasets, and six attack methods, sweeping over 336 distinct attack settings. Through it, we first challenge the prevailing notion of MM conferring free adversarial robustness, and show {that} MM cannot reliably defend against transfer attacks, with {over $80\%$ transfer rate}. Moreover, we reveal {two} key insights for machine-learning practitioners regarding MM and transferability for a robust system design: (1) stronger MM methods increase vulnerability to transfer attacks and (2) mitigating representation bias increases vulnerability to transfer attacks. {We also find weight averaging as an exception to (1), i.e., it is found to be the most vulnerable method to transfer attacks, despite being the weakest MM method.  Finally, we analyze the underlying reasons for {these findings}, show how an adversary with limited information could launch an attack, and provide potential solutions.}  {These findings offer actionable insights for deploying MM in security-sensitive machine-learning systems.}

\end{abstract}

\section{Introduction}
\label{section:introduction}

Pre-Trained Models~(PTMs) are becoming increasingly popular across various domains, ranging from image recognition to natural language processing~\cite{bert,jia2021scaling,clip}. They offer powerful representations that can be adapted to various downstream tasks through fine-tuning. However, fine-tuning often leads to degradation of multi-task performance, i.e., tasks other than the one being optimized~\cite{castastrophic_forgetting}. To counteract this issue, Model Merging~(MM)~\cite{model_merge_average,task_arithmetic,ties_merging,ada_merging,representation_surgery} has recently been shown as an effective alternative, with over 40,000 merged models on HuggingFace\footnote{ \hyperref{https://huggingface.co/models?other=merge}{}{}{https://huggingface.co/models?other=merge}}. MM enables combining multiple fine-tuned models
into a single model that retains or even enhances performance across multiple tasks, without training or access to task-specific data. 


Despite its growing popularity, the security of MM methods remains largely unexplored. Recent research has shown that MM can mitigate {various} types of adversarial attacks~\cite{free_lunch,zhou2025mixup,gangwal2025poster,hiromi2025enhancing,yang2024model,wang2024rethinking,farn2024safeguard}. {For example}, Arora et al.~\cite{free_lunch} demonstrated that merging a backdoored model with benign models can neutralize the backdoor, effectively sanitizing the merged model and appearing to provide a ``free lunch'' of adversarial robustness. {However, the field of adversarial robustness has also repeated encountered a pattern, where heuristic defenses that offer no formal guarantees do not hold up over time~\cite{athalye2018obfuscated}. In this work, we explore a similar pattern, and show that} while MM may mitigate {certain kinds of adversarial attacks, like backdoor attacks}, it actually increases vulnerability to another class of adversarial threats: (transferable) adversarial examples, which are subtly modified inputs that can mislead models into making incorrect predictions~\cite{szegedy2013intriguing}. These examples also exhibit the property of \textit{transferability}~\cite{adversarial_transferability}, i.e., the tendency to remain effective even on models trained with different datasets or with different architecture, so long as both models perform the same task. This enables an adversary to launch a \textit{transfer attack}, where an adversarial example crafted on a surrogate model can successfully mislead a target model, even without access to its parameters or architecture~\cite{papernot2016transferability}.

\begin{figure}[!htbp]
    \centering
    \includegraphics[width=\linewidth]{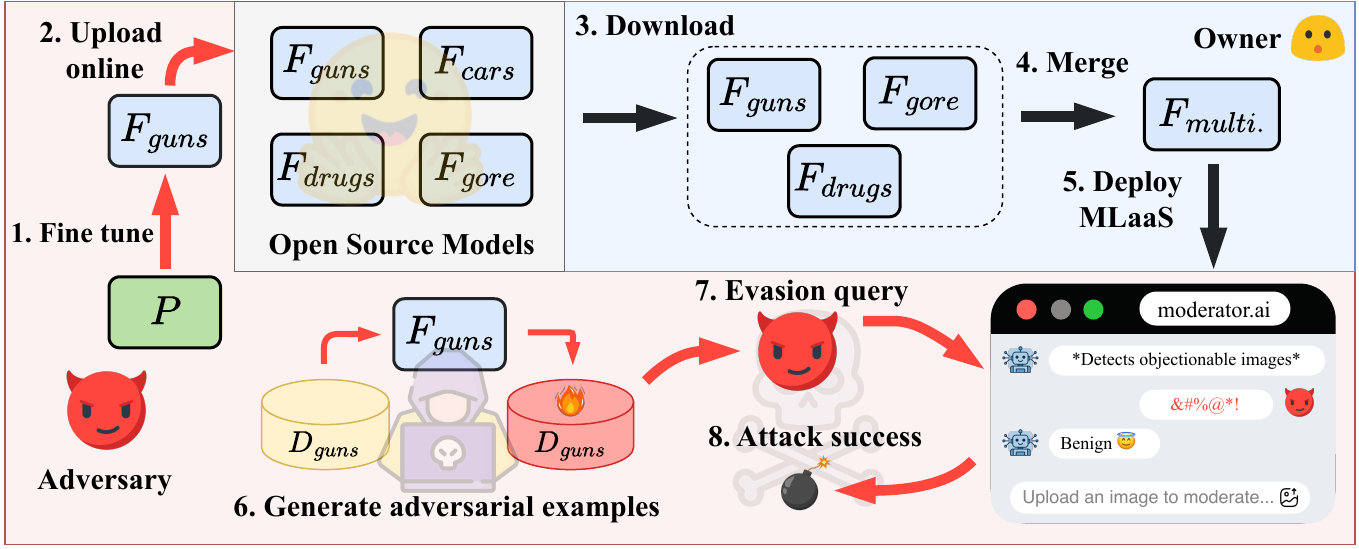}
    \caption{Illustration of a possible attack strategy of adversary. (1) Adversary fine-tunes the PTM for classifying objectionable content (Guns) and (2) uploads it online on open-source platforms like HuggingFace. (3) The owner~(victim) downloads multiple models fine-tuned to classify objectionable contents~(e.g., guns, drugs, gore) and (4) merges them into a single model. (5) Owner then deploys the merged model as an MLaaS. (6) Adversary then generates adversarial examples using either the previously uploaded fine-tune model or off-the-shelf PTM as surrogate and (7) launches an evasion attack on owner's MLaaS. Finally, due to architectural or decision boundary similarity between surrogate and target model, (8) the attack succeeds.}
    \label{fig:adv-trans}
\end{figure}

Existing literature on adversarial transferability suggests {as two models become accurate, the transferability rate between them increases~\cite{ebrahimpourtraining}. This implies weaker~(i.e., less accurate) MM methods, like weight averaging, could mitigate transfer attacks. However, it is also known that adversarial examples} tend to transfer well between models having same architecture~\cite{higher_test_transferability1}. Therefore, in order to evade a Machine Learning as a Service~(MLaaS) system, {e.g., an image content moderation system}, that internally employs a merged model, an adversary can use as the surrogate either (1) the publicly available PTM that shares the same architecture with the merged model, or (2) one of the popular fine-tuned models~{(highly accurate for the specific task)} that was used in the merged model's MM process, and expect the transfer attack to succeed with high probability. Figure~\ref{fig:adv-trans} illustrates a possible attack strategy of adversary.

Beyond this apparent threat, it also remains unclear whether certain MM methods are particularly vulnerable to transfer attacks, {or how mitigating representation bias~\cite{representation_surgery} influences transferability}. 
To our best knowledge, none of the existing works have extensively studied the relation between transferable adversarial examples and MM. Therefore, we study MM under this threat model, and find several statistically validated empirical results that challenge the status
quo. We further give explanations for our findings and discuss potential solutions to address this problem.

\par
\textit{Contribution:} The major contributions of our work are: 
\begin{enumerate}
	\setlength{\itemsep}{0cm}
	\setlength{\parskip}{0cm}
	\item We {first} challenge the prevailing notion of MM conferring free adversarial robustness, by showing it is vulnerable to transfer attacks, {with a transfer rate exceeding $80\%$~(cf. \S\ref{section:evaluation}).}
	\item We statistically validate our empirical results in more than 336 different attack settings~(eight MM methods, seven datasets, six attack methods) {show how}:
    \begin{enumerate}
        \item Stronger MM method increases the risk of transfer attacks~(cf. \S\ref{subsection:hypothesis1}).
        \item Reducing representation bias increases the risk of transfer attacks~(cf. \S\ref{subsection:hypothesis2}).
        \item {Weight averaging is most vulnerable to transfer attacks, despite being the weakest MM methods~(cf. \S\ref{subsection:hypothesis3}). }
    \end{enumerate}
    \item {We explain the reasons for each of our findings~(\S\ref{subsection:hyp1_hyp2_discussion}, \S\ref{subsection:hyp3_discussion}), demonstrate how an adversary with limited information could realistically launch a transfer attack~(\S\ref{subsection:limited_information_attacks}), and give a potential solution for robust MM~(\S\ref{subsection:solution_and_conjecture}).} 
    In the spirit of reproducible research, we provide our codebase\footnote{Code available at: {\hyperref{https://github.com/aaryanajaysharma/merge-now-regret-later}{}{}{https://github.com/aaryanajaysharma/merge-now-regret-later}}}.
\end{enumerate}
\textit{Organization:} The remainder of this paper is organized as follows. \S\ref{section:background}  presents the relevant background knowledge for {the} paper. We elucidate our threat model in \S\ref{section:threatmodel}. \S\ref{section:experiment_details} details our experimental setup, followed by evaluation results in \S\ref{section:evaluation}. \S\ref{section:related_work} summarizes related works. \S\ref{section:discussion} discusses explanation of findings, solutions and other results, and \S\ref{section:conclusion} concludes.
\section{Background}
\label{section:background}
\textbf{Notations.} Let $f_{\boldsymbol{\theta}}: \mathbb{R}^D \rightarrow [0, 1]^c$ denote an image classification neural network model parameterized by the vector $\boldsymbol{\theta} \in \mathbb{R}^n$. The input is a vector $\mathbf{x}_i \in \mathbb{R}^D$, and the model outputs a probability distribution over $c$ classes: $\hat{\mathbf{y}}_i = f_{\boldsymbol{\theta}}(\mathbf{x}_i) = [\hat{y}_{i,1}, \ldots, \hat{y}_{i,c}]^\top \in [0, 1]^c$, with $\sum_{j=1}^c \hat{y}_{i,j} = 1$. Given a dataset $\mathcal{D}_{\mathit{tr}} = \{(\mathbf{x}_i, \mathbf{y}_i)\}_{i=1}^{N_{tr}}$, the ground truth label $\mathbf{y}_i \in \{0,1\}^c$ is represented in one-hot encoding. The model predicts label (class index) $\arg\max_j \hat{y}_{i,j}$. Here, $\mathit{n}$ denotes the number of parameters, $\mathit{D}$ is the input dimension, $N_{tr}=|\mathcal{D}_{\mathit{tr}}|$ is the number of training samples, and $\mathit{c}$ is the number of output classes. \tablename~\ref{tab:notations} summarizes the notation used in our work.


\subsection{Model Merging}
\label{subsection:model_merging}
Consider $\mathit{T}$ independently fine-tuned models $\{f_{\boldsymbol{\theta}_t}\}_{t=1}^T$, each trained on its own, task-specific dataset $\mathcal{D}^t_{\mathit{tr}} \subseteq \mathbb{R}^D \times \{0,1\}^c$. Following prior work~\cite{representation_surgery}, we assume all $f_{\boldsymbol{\theta}_t}$ are fine-tuned from a common pre-trained backbone $f_{\boldsymbol{\theta}_0}$ (e.g., ViT~\cite{vit}, ResNet~\cite{resnet}). MM aims to combine the parameter vectors $\{\boldsymbol{\theta}_t\}_{t=1}^T$ of $T$ independently fine-tuned models into a single parameter vector $\boldsymbol{\theta}_{\mathit{mtl}}^{m}$, such that the resulting model $f_{\boldsymbol{\theta}_{\mathit{mtl}}^m}$ can simultaneously perform well on all tasks $t \in [T]$. Mathematically, $\zeta_m: \mathbb{R}^{n \times T} \rightarrow \mathbb{R}^{n}$ such that $ \zeta_m(\boldsymbol{\Theta})=\boldsymbol{\theta}_{\mathit{mtl}}^{m}$, where $\boldsymbol{\Theta} = [\boldsymbol{\theta}_1 \; \boldsymbol{\theta}_2 \; \cdots \; \boldsymbol{\theta}_T] \in \mathbb{R}^{n \times T}$ is combined parameter matrix, $\zeta_m$ is the function performing MM and $m$ denotes a specific merging method. 

A key constraint for MM is that access to the task-specific fine-tuning data $\{\mathcal{D}_{\mathit{tr}}^t\}_{t=1}^T$ is restricted during the merging process. The objective therefore, is to minimize the average loss of the merged model across all task-specific test sets $\{\mathcal{D}_{\mathit{te}}^t\}_{t=1}^T$, where each test set contains labeled pairs $(\mathbf{x}_i, \mathbf{y}_i) \in \mathbb{R}^D \times \{0,1\}^c$. Formally, the optimization goal is:
\begin{equation}
    \label{eq:model_merging_goal}
    \min_{\boldsymbol{\theta}_{\mathit{mtl}}^m} \frac{1}{T} \sum_{t=1}^T \frac{1}{|\mathcal{D}_{\mathit{te}}^t|} \sum_{i=1}^{|\mathcal{D}_{\mathit{te}}^t|} \mathcal{L}(f_{\boldsymbol{\theta}_{\mathit{mtl}}^m}(\mathbf{x}_i), \mathbf{y}_i),
\end{equation}
where $\mathcal{L}(\cdot)$ is a task-specific loss function, such as the cross-entropy loss.

We discuss four representative \textit{pre-hoc} MM methods (i.e., operating to obtain $\boldsymbol{\theta}_{\mathit{mtl}}^{m}$) and one \textit{post-hoc} MM method (i.e., operating after $\boldsymbol{\theta}_{\mathit{mtl}}^{m}$ is obtained) considered in our work below:

\begin{enumerate}
    \item {\textbf{Pre-hoc merging methods.}
    All pre-hoc methods operate on $T$ fine-tuned models $\{\boldsymbol{\theta}_t\}_{t=1}^T$ that share a common pretrained initialization
    $\boldsymbol{\theta}_0 = \{\boldsymbol{\theta}_0^{\ell}\}_{\ell=1}^L$.
    For each task $t$ and layer $\ell \in [L]$, where $L$ denotes total number of layers, we define a layer-wise task vector}
    $\boldsymbol{\tau}_t^{\ell} \triangleq \boldsymbol{\theta}_t^{\ell} - \boldsymbol{\theta}_0^{\ell}.$
    {A generic pre-hoc MM method $m$ produces merged parameters}
    \begin{equation}
    \label{eq:generic_pre_hoc}
    \boldsymbol{\theta}_{\mathit{mtl}}^{m}
    =
    \Bigl\{
        \boldsymbol{\theta}_0^{\ell}
        +
        \sum_{t=1}^T
            \lambda_{t}^{\ell,m}\,
            \Psi_m\bigl(\{\boldsymbol{\tau}_u^{\ell}\}_{u=1}^T\bigr)_t
    \Bigr\}_{\ell=1}^L,
\end{equation}
{where $\Psi_m$ is a method-specific transformation of the layer-wise task vectors and
$\lambda_{t}^{\ell,m} \in \mathbb{R}$ are scaling coefficients that may depend on both $t$ and $\ell$.}

{Equation~\eqref{eq:generic_pre_hoc} covers the four pre-hoc methods in our study:
(i) {Weight Averaging~(WA)~\cite{model_merge_average}} uses the identity transform $\Psi_{\text{WA}}$ and uniform coefficients $\lambda_{t}^{\ell,\text{WA}} = 1/T$ for all $t,\ell$;
(ii) {Task Arithmetic~(TA)~\cite{task_arithmetic}} uses the identity transform with a single global scalar $\lambda_{t}^{\ell,\text{TA}} \equiv \lambda$;
(iii) {TIES-Merging~(TM)~\cite{ties_merging}} applies the TIES~(\texttt{TRIM}, \texttt{ELECT SIGN}, and \texttt{MERGE}) operations inside $\Psi_{\text{TM}}$ and also uses a global scalar $\lambda_{t}^{\ell,\text{TM}} \equiv \lambda$;
(iv) {AdaMerging~(AM)~\cite{ada_merging}} uses the identity transform and learns the coefficients, either task-wise ($\lambda_{t}^{\ell,\text{AM}} = \lambda_t$ for all $\ell$) or layer-wise (free $\lambda_{t}^{\ell,\text{AM}}$). 
}

    \item \textbf{Representation Surgery~(RS)~\cite{representation_surgery}.} RS is a post-hoc MM method, in the sense that it does not combine $\{\boldsymbol{\theta}_t\}_{t=1}^T$ into $\boldsymbol{\theta}^m_{mtl}$. Rather, it first observes that both simple and advanced MM techniques often suffer from a \textit{representation bias}---a consistent, task-specific discrepancy between the feature representations of the merged model and those of the individual fine-tuned models, and tries to mitigate this representation bias. To achieve this, RS introduces a task-specific adapter function ${\Phi}_t(\cdot)$ for each task $t \in [T]$, which aims to correct for the task-specific bias in the representation space. Let $\boldsymbol{Z}_{\mathit{mtl}}^t \in \mathbb{R}^{|\mathcal{D}_{\mathit{te}}^t| \times k}$ and $\boldsymbol{Z}_{\mathit{ind}}^t \in \mathbb{R}^{|\mathcal{D}_{\mathit{te}}^t| \times k}$ denote the feature matrices extracted from the merged model and the individual fine-tuned model respectively for task $t$, where $k$ is the feature dimension. Each adapter is parameterized by $\boldsymbol{\theta}_{\Phi_t}$ and learns to estimate the representation bias via the transformation $\Phi_t(\boldsymbol{Z}_{\mathit{mtl}}^t)$. The goal is to minimize the average $\ell_1$ distance between the feature representations of the merged model and the corresponding individual model: \begin{equation} \label{eq:representation_surgery} \arg\min_{\{\boldsymbol{\theta}_{\Phi_t}\}_{t=1}^T} \;\frac{1}{T} \sum_{t=1}^T \frac{1}{|\mathcal{D}_{\mathit{te}}^t|} \left\| \widehat{\boldsymbol{Z}}_{\mathit{mtl}}^t - \boldsymbol{Z}_{\mathit{ind}}^t \right\|_1, \end{equation} subject to the constraint: 
$\widehat{\boldsymbol{Z}}_{\mathit{mtl}}^t = \boldsymbol{Z}_{\mathit{mtl}}^t - \Phi_t(\boldsymbol{Z}_{\mathit{mtl}}^t), $ where $\widehat{\boldsymbol{Z}}_{\mathit{mtl}}^t$ denotes the debiased representation for task $t$. Due to this post-hoc removal of representation bias, RS can be combined with any \textit{pre-hoc} MM method. 
\end{enumerate}

Following our previous discussion, let $\mathcal{M}$ denote the space of MM methods considered in this work. Specifically, $\mathcal{M}$ is defined as the Cartesian product of two sets: $\{\text{WA, TA, TM, AM}\}$, denoting four representative merging methods, and $\{\varnothing,\, \text{RS}\}$, indicating the absence or presence of RS, respectively, i.e., $\mathcal{M} = \{\text{WA, TA, TM, AM}\} \times \{\varnothing, \,\text{RS}\}.$ Each merging method $m$ considered in our study is an element of this space, i.e., $m \in \mathcal{M}$. For ease of notation, we shall refer to the method $m = (u, \,\varnothing)$ as simply $u$ and $m = (u, \,\text{RS})$ as $u+\text{RS}$ henceforth (e.g., $(\text{WA}, \,\varnothing) \mapsto \text{WA}$, $(\text{WA}, \,\text{RS}) \mapsto \text{WA+RS}$).

\subsection{Adversarial Examples}
\label{subsection:adversarial_examples}
Adversarial examples are perturbed inputs intentionally crafted to cause a trained neural network model to produce incorrect predictions, despite the perturbation being imperceptibly small to humans~\cite{szegedy2013intriguing}. Formally, a valid adversarial example is a perturbed input $\tilde{\mathbf{x}}_i = \mathbf{x}_i + \boldsymbol{\delta}_i$ that satisfies the following properties~\cite{papernot2017practical,carlini2019evaluating}:
\begin{itemize}
    \item \textbf{Bounded Perturbation.} The perturbation vector $\boldsymbol{\delta}_i \in \mathbb{R}^D$ is constrained such that $\|\boldsymbol{\delta}_i\|_p \leq \epsilon$ where $\epsilon > 0$ is the perturbation budget and norm $\|\cdot\|_p$ (e.g., $\ell_\infty$, $\ell_2$).
    
    \item \textbf{Prediction Change.} The model’s predicted class on the adversarial example differs from the original prediction, i.e.,
    \begin{equation}
    \label{eq:adv_def}
        \arg\max_j f_{\boldsymbol{\theta}}(\tilde{\mathbf{x}}_i)_j \neq \arg\max_j f_{\boldsymbol{\theta}}(\mathbf{x}_i)_j.
    \end{equation}
\end{itemize}

One of the surprising properties of adversarial examples is their \textit{transferability}~\cite{papernot2016transferability}---the phenomenon where an adversarial example $\tilde{\mathbf{x}}_i$ crafted to fool one model $f_{\boldsymbol{\theta}}$ can also fool a different model $f_{\boldsymbol{\theta}'}$, even if the second model has a different architecture, training data, or initialization, as long as they perform the same task. Formally, we say an adversarial example transfers from $f_{\boldsymbol{\theta}}$ to $f_{\boldsymbol{\theta}'}$, if it satisfies both Eq.~\eqref{eq:adv_def} and Eq.~\eqref{eq:adv_transfer_def} as follows:
\begin{equation}
\label{eq:adv_transfer_def}
    \arg\max_j f_{\boldsymbol{\theta}'}(\tilde{\mathbf{x}}_i)_j \neq \arg\max_j f_{\boldsymbol{\theta}'}(\mathbf{x}_i)_j.
\end{equation}

Transferability enables the feasibility of (black-box) transfer attacks, where an adversary crafts adversarial examples using a surrogate model $f_{\boldsymbol{\theta}}$ and launches them against a target model $f_{\boldsymbol{\theta}'}$ for evasion, without having access to its parameters or gradients. Several factors influence transferability, including model architecture similarity~\cite{tramer2017space}, alignment of decision boundaries~\cite{liu2017delving}, and the choice of attack method~\cite{adversarial_transferability}.


Various attack methods to generate adversarial examples have been proposed by optimizing the perturbation $\boldsymbol{\delta}_i$ under norm constraints. We use the following five first-order attacks: (1) Fast Gradient Sign Method~(FGSM)~\cite{goodfellow2014explaining}; (2) Iterative FGSM~(I-FGSM)~\cite{kurakin2018adversarial}; (3) Projected Gradient Descent~(PGD)~\cite{madry2018towards}; (4) Nesterov I-FGSM~(NI-FGSM)~\cite{ni_fgsm}; and (5) Translation-Invariant FGSM~(TI-FGSM)~\cite{ti_fgsm}; {We also use one query-based black-box attack, Square Attack~\cite{andriushchenko2020square}, in our work.}  

Consequently, we define the space of attack methods $\Gamma$ as $\{\text{FGSM, I-FGSM, } $ $\text{ PGD, NI-FGSM, TI-FGSM}\}$\footnote{We exclude Square attack from $\Gamma$ for notational convenience and uniformity when performing statistical analysis in \S\ref{section:evaluation}.}, with attack method $\gamma$ being an element of this space, i.e., $\gamma \in \Gamma$.

\section{Threat Model}
\label{section:threatmodel}

\textbf{Actors.} We have two actors in our threat model: (1) owner \owner, who uses MM method $m \in \mathcal{M}$ for creating model  $f_{\boldsymbol{\theta}_{\mathit{mtl}}^m}$. \owner employs $f_{\boldsymbol{\theta}_{\mathit{mtl}}^m}$ in her system and makes it publicly available as an MLaaS via an API; and (2) adversary \adversary, which has illicit gains for evading/bypassing \owner's system.

\textbf{Adversary's Goal.}
\adversary's goal is make \owner's model $f_{\boldsymbol{\theta}_{\mathit{mtl}}^m}$ change its prediction on an input of interest $\mathbf{x}$. Formally, \adversary's goal is finding $\tilde{\mathbf{x}}=\mathbf{x}+\boldsymbol{\delta}$ such that: $\arg\max_j f_{\boldsymbol{\theta}_{\mathit{mtl}}^m}(\tilde{\mathbf{x}})_j \neq \arg\max_j f_{\boldsymbol{\theta}_{\mathit{mtl}}^m}(\mathbf{x})_j,$
with $|\boldsymbol{\delta}|_{p}\leq \epsilon$. {As a concrete example, \owner's MLaaS system can be an automatic image moderation service like PicPurify\footnote{\href{https://www.picpurify.com/demo-gun.html}{https://www.picpurify.com/demo-gun.html}} or sightengine\footnote{\href{https://sightengine.com/detect-weapons-alcohol-drugs}{https://sightengine.com/detect-weapons-alcohol-drugs}}, where \adversary's gains from evading such moderation system.}

\textbf{Adversary's Capabilities.}
\label{subsection:adversary_capabilities}
Following prior works on image-classification black-box adversarial attacks~\cite{ti_fgsm,ni_fgsm}, we consider an $\ell_{\infty}$-norm constrained threat model, i.e, $|\tilde{\mathbf{x}}-\mathbf{x}|_{\infty} \leq \epsilon$. This implies \adversary is capable of perturbing any pixel of the image by value at most $\epsilon$. Unless otherwise stated, we assume \adversary performs untargeted transfer attack since they are harder to defend against, and since targeted transfer attacks, although practical, are much harder to perform~\cite{li2020towards}. We discuss the results on targeted attacks in \S\ref{subsection:targeted_attacks}.

\textbf{Adversary's Knowledge.}
We make three {assumptions} regarding \adversary's knowledge: (1) \adversary knows \owner has used MM while creating her model; (2) \adversary knows the specific architecture of the model being merged; (3) \adversary has access to one of the two kinds of surrogate model (that is used by the \owner while creating her merged model): (a) PTM $f_{\boldsymbol{\theta}_{\mathit{0}}}$ or (b) fine-tuned model $f_{\boldsymbol{\theta}_{\mathit{t}}}$ on task $t$, which she uses to craft transferable adversarial examples using attack method 
$\gamma \in \Gamma$. Assumption (3) is realistic in the scenario where \owner builds her merged model by downloading off-the-shelf fine-tuned models, as illustrated in \figurename~\ref{fig:adv-trans}. {We also relax few of these assumptions, i.e., where \adversary's knowledge is limited while performing attack in \S\ref{subsection:limited_information_attacks}}. Following previous works on MM~\cite{ada_merging,representation_surgery}, we assume \adversary, like \owner, does not have access to any task-specific training data $\mathcal{D}^t_{\mathit{tr}}$, while having partial access to \textit{unlabeled} task-specific test data $\mathcal{D}^t_{\mathit{te}}$ {and label space for attack optimization}.

\section{Experimental Details}
\label{section:experiment_details}
\textbf{Experimental Setup.} We run all our experiments on an NVIDIA DGX A100 machine using the PyTorch framework, and use Adam as the optimizer.
\par
\textbf{Datasets and Models.} To capture diverse domains, we selected 7 image datasets: ($t=1$) Cars~\cite{cars}, ($t=2$) MNIST~\cite{mnist}, ($t=3$) EuroSAT~\cite{eurosat}, ($t=4$) GTSRB~\cite{gtsrb}, ($t=5$) SVHN~\cite{svhn}, ($t=6$) RESISC45~\cite{resisc} and ($t=7$) DTD~\cite{dtd}, in accordance with previous work on MM~\cite{representation_surgery}. Please refer to \appendixname~\ref{subsection:dataset} for dataset description. Following prior works~\cite{representation_surgery}, we mainly use a CLIP model~\cite{clip} with ViT-32/B~\cite{vit} as the backbone for classification. We also use a CLIP model with ResNet-50~\cite{he2016deep} backbone in \S\ref{subsection:limited_information_attacks}. Model accuracies for each model considered in our work is given in \tablename~\ref{tab:performance_vitbase32}~(\appendixname~\ref{section:experimentalsetting_appendix}).

\begin{table}[htbp]
  \centering
  \caption{{Transfer Rate matrices ($\{\mathbf{R}_\gamma^t\}_{t=1}^T$) for attack method $\gamma = \text{NI-FGSM}$ (in \%), showing transferability from surrogate model $s \in \mathcal{S} = \{f_{\boldsymbol{\theta}_0}, f_{\boldsymbol{\theta}_t}\}$ to target models $\mathcal{T}$.  $\{\overline{R}_{\gamma}^{t,\,s}\}_{t=1}^T$ is mean Transfer Rate.} }
  \resizebox{\linewidth}{!}{
    \begin{tabular}{l|c|cc|cccccccc|c}
    \toprule
     Task $t$ & $\mathcal{S}$ $\downarrow$ / $\mathcal{T}$ $\rightarrow$ & $f_{\boldsymbol{\theta}_0}$ & $f_{\boldsymbol{\theta}_t}$ & $f_{\boldsymbol{\theta}_{mtl}^{\text{WA}}}$ & $f_{\boldsymbol{\theta}_{mtl}^{\text{TA}}}$ & $f_{\boldsymbol{\theta}_{mtl}^{\text{TM}}}$ & $f_{\boldsymbol{\theta}_{mtl}^{\text{AM}}}$ & $f_{\boldsymbol{\theta}_{mtl}^{\text{WA+RS}}}$ & $f_{\boldsymbol{\theta}_{mtl}^{\text{TA+RS}}}$ & $f_{\boldsymbol{\theta}_{mtl}^{\text{TM+RS}}}$ & $f_{\boldsymbol{\theta}_{mtl}^{\text{AM+RS}}}$ & $\overline{R}_{\gamma}^{t,\,s}$ \\
    \midrule
    \multirow{2}{*}{\centering $(t=1)$ Cars} & $s = f_{\boldsymbol{\theta}_0}$ & \gradient{100.00} & \gradient{87.32} & \gradient{92.11} & \gradient{82.26} & \gradient{87.54} & \gradient{90.98} & \gradient{91.77} & \gradient{80.42} & \gradient{86.86} & \gradient{89.13} & \gradient{87.63} \\
     & $s = f_{\boldsymbol{\theta}_t}$ & \gradient{83.62} & \gradient{100.00} & \gradient{98.58} & \gradient{97.64} & \gradient{98.43} & \gradient{98.35} & \gradient{98.45} & \gradient{97.28} & \gradient{98.50} & \gradient{98.27} & \gradient{98.19} \\
    \midrule
    \multirow{2}{*}{\centering $(t=2)$ MNIST} & $s = f_{\boldsymbol{\theta}_0}$ & \gradient{100.00} & \gradient{21.21} & \gradient{72.09} & \gradient{44.85} & \gradient{58.69} & \gradient{36.53} & \gradient{70.47} & \gradient{40.52} & \gradient{62.00} & \gradient{38.17} & \gradient{52.91} \\
     & $s = f_{\boldsymbol{\theta}_t}$ & \gradient{93.23} & \gradient{100.00} & \gradient{98.92} & \gradient{99.34} & \gradient{99.22} & \gradient{99.86} & \gradient{99.42} & \gradient{99.64} & \gradient{99.62} & \gradient{99.86} & \gradient{99.48} \\
    \midrule
    \multirow{2}{*}{\centering $(t=3)$ EuroSAT} & $s = f_{\boldsymbol{\theta}_0}$ & \gradient{100.00} & \gradient{67.38} & \gradient{79.96} & \gradient{78.83} & \gradient{76.27} & \gradient{74.24} & \gradient{72.39} & \gradient{60.47} & \gradient{62.79} & \gradient{66.55} & \gradient{71.44} \\
     & $s = f_{\boldsymbol{\theta}_t}$ & \gradient{56.21} & \gradient{100.00} & \gradient{91.74} & \gradient{90.22} & \gradient{92.00} & \gradient{96.15} & \gradient{98.56} & \gradient{98.70} & \gradient{98.93} & \gradient{98.70} & \gradient{95.62} \\
    \midrule
    \multirow{2}{*}{\centering $(t=4)$ GTSRB} & $s = f_{\boldsymbol{\theta}_0}$ & \gradient{100.00} & \gradient{31.08} & \gradient{75.50} & \gradient{59.73} & \gradient{64.92} & \gradient{64.03} & \gradient{66.13} & \gradient{52.43} & \gradient{58.59} & \gradient{59.30} & \gradient{62.58} \\
     & $s = f_{\boldsymbol{\theta}_t}$ & \gradient{82.62} & \gradient{100.00} & \gradient{95.77} & \gradient{96.28} & \gradient{97.02} & \gradient{95.98} & \gradient{97.42} & \gradient{96.68} & \gradient{97.78} & \gradient{96.89} & \gradient{96.73} \\
    \midrule
    \multirow{2}{*}{\centering $(t=5)$ SVHN} & $s = f_{\boldsymbol{\theta}_0}$ & \gradient{100.00} & \gradient{20.95} & \gradient{66.40} & \gradient{56.95} & \gradient{58.57} & \gradient{52.77} & \gradient{61.01} & \gradient{49.91} & \gradient{52.76} & \gradient{55.81} & \gradient{56.77} \\
     & $s = f_{\boldsymbol{\theta}_t}$ & \gradient{99.36} & \gradient{100.00} & \gradient{91.01} & \gradient{94.41} & \gradient{94.69} & \gradient{95.54} & \gradient{91.97} & \gradient{94.77} & \gradient{95.30} & \gradient{96.36} & \gradient{94.26} \\
    \midrule
    \multirow{2}{*}{\centering $(t=6)$ RESISC45} & $s = f_{\boldsymbol{\theta}_0}$ & \gradient{100.00} & \gradient{70.40} & \gradient{79.05} & \gradient{71.37} & \gradient{74.77} & \gradient{74.43} & \gradient{79.09} & \gradient{71.24} & \gradient{72.12} & \gradient{72.17} & \gradient{74.28} \\
     & $s = f_{\boldsymbol{\theta}_t}$ & \gradient{71.99} & \gradient{100.00} & \gradient{96.46} & \gradient{95.92} & \gradient{96.85} & \gradient{97.27} & \gradient{96.43} & \gradient{96.98} & \gradient{98.26} & \gradient{98.23} & \gradient{97.05} \\
    \midrule
    \multirow{2}{*}{\centering $(t=7)$ DTD} & $s = f_{\boldsymbol{\theta}_0}$ & \gradient{100.00} & \gradient{52.77} & \gradient{66.78} & \gradient{53.42} & \gradient{58.96} & \gradient{63.52} & \gradient{65.15} & \gradient{54.72} & \gradient{58.31} & \gradient{59.28} & \gradient{60.02} \\
     & $s = f_{\boldsymbol{\theta}_t}$ & \gradient{22.16} & \gradient{100.00} & \gradient{88.02} & \gradient{85.15} & \gradient{87.07} & \gradient{87.07} & \gradient{92.22} & \gradient{87.31} & \gradient{91.74} & \gradient{92.10} & \gradient{88.83} \\
    \bottomrule
    \end{tabular}
  }
  \label{tab:tr_matrix_nifgsm}
\end{table}

\textbf{Attack Methods.} For attack methods, we employ the standard FGSM~\cite{goodfellow2014explaining} and the state-of-the-art method: NI-FGSM~\cite{ni_fgsm} and TI-FGSM~\cite{ti_fgsm}, designed for black-box transfer settings. We also consider I-FGSM~\cite{kurakin2018adversarial} and PGD~\cite{madry2018towards}, which are generally known to perform better in white-box settings, to provide additional context\footnote{{We exclude Carlini \& Wagner~(C\&W) attack~\cite{carlini2017towards}, as it is primarily designed for white-box setting~\cite{boosting_ae1}.}}. Finally, to evaluate MM against query-based black-box attack, we employ Square Attack~\cite{andriushchenko2020square}. We split the test data into two disjoint halves, and use one half for evaluation of merged models and the other to generate adversarial examples. Please refer to \appendixname~\ref{subsection:hyperparameters} for attack hyperparameters.
\par
\textbf{Metric.} We use the ASR to measure the success of a transfer attack. ASR is defined as the proportion of test examples for which the top-1 prediction of the target model $f_{\boldsymbol{\theta}'}$ on the adversarial input $\tilde{\mathbf{x}}_i$ differs from its prediction on the corresponding clean input $\mathbf{x}_i$~\cite{papernot2016transferability}. Mathematically, we define ASR as:
{\small
\begin{align}
\label{eq:asr}
\frac{1}{|\mathcal{D}_{\mathit{te}}^t|} \sum_{i=1}^{|\mathcal{D}_{\mathit{te}}^t|} \mathbb{I}\bigl(\arg\max_j f_{\boldsymbol{\theta}'}(\tilde{\mathbf{x}}_i)_j \neq \arg\max_j f_{\boldsymbol{\theta}'}(\mathbf{x}_i)_j\bigr),
\end{align}
}
 where $\mathbb{I}(\cdot)$ is the indicator function. {While defining ASR, an additional constraint is also typically imposed, where only those samples which are correctly classified by the model are considered~\cite{kurakin2018adversarial}. However, since both \owner and \adversary do not have access to labeled test dataset in MM setting~(\cf \S\ref{section:threatmodel}), we focus on ASR induced by prediction change~\cite{carlini2019evaluating,papernot2016transferability,moosavi2017universal}. }

\section{Evaluating Transferability in Model Merging}
\label{section:evaluation}
We first aim to find the answer to the following question:
\begin{tcolorbox}
    \textbf{How Successful is Model Merging in Defending Against Transfer Attacks?} \textit{Specifically, to what extent does the success of \adversary in transfer attacks differ from that in the white-box scenario?}
\end{tcolorbox}
{To answer this, we follow existing literature~\cite{tramer2017space,ebrahimpourtraining} and define the \textit{transfer rate} between the surrogate and target model as follows.}

{Let $\mathcal{D}_{\mathit{te}}^t = \{\mathbf{x}_i^t\}_{i=1}^{n_t}$ denote the test set for task $t$, and let $\mathbf{x}_{i,\gamma}^{t,s}$ be the adversarial example for $\mathbf{x}_i^t$ crafted against surrogate $s$ with attack $\gamma$. For each surrogate $s \in \mathcal{S}=\{f_{\boldsymbol{\theta}_0}, f_{\boldsymbol{\theta}_t}\}$ and target $l \in \mathcal{T} = \mathcal{S} \cup \{f_{\boldsymbol{\theta}_{\mathit{mtl}}^m} \mid m \in \mathcal{M}\}$\footnote{{We include $\mathcal{S}$ in $\mathcal{T}$ for completeness of cross-evaluation.}} we define}
\begin{align}
\mathcal{X}_{\gamma}^{t,s} &= \bigl\{ \mathbf{x}_{i,\gamma}^{t,s} \mid i \in [n_t],\, f_s(\mathbf{x}_{i,\gamma}^{t,s}) \neq f_s(\mathbf{x}_i^t) \bigr\}, \\
\mathcal{Z}_{\gamma}^{t,s\rightarrow l} &= \bigl\{ \mathbf{x}_{i,\gamma}^{t,s} \mid i \in [n_t],\, f_l(\mathbf{x}_{i,\gamma}^{t,s}) \neq f_l(\mathbf{x}_i^t) \bigr\},
\end{align}
{where $\mathcal{X}_{\gamma}^{t,s}$ is the set of adversarial test examples on which the surrogate changes its top-1 prediction under attack and $\mathcal{Z}_{\gamma}^{t,s\rightarrow l}$ is the set of adversarial test examples on which the target changes its top-1 prediction under the same transferred adversarial examples. We then define the \emph{transfer rate matrix} $\mathbf{R}_\gamma^t \in [0, 1]^{|\mathcal{S}| \times |\mathcal{T}|}$ with entries}
\begin{equation}
[\mathbf{R}_\gamma^t]_{s,\,l} =
\begin{cases}
\dfrac{\bigl|\mathcal{X}_{\gamma}^{t,s} \cap \mathcal{Z}_{\gamma}^{t,s\rightarrow l}\bigr|}{\bigl|\mathcal{X}_{\gamma}^{t,s}\bigr|}, & \text{if } \bigl|\mathcal{X}_{\gamma}^{t,s}\bigr| > 0,\\[0.75em]
0, & \text{otherwise,}
\end{cases}
\end{equation}
{which estimates the conditional probability that the transfer attack also changes the top-1 prediction of target model $l$ given that it has changed the top-1 prediction of surrogate model $s$.}

{To quantify the average success of transfer attack relative to its success in surrogate, we take the average of transfer rates over target models. Mathematically,}
\begin{equation}
\overline{R}_{\gamma}^{t,\,s}=\frac{1}{|\mathcal{T}\backslash\mathcal{S}|}\sum^{}_{l\in \mathcal{T} \backslash \mathcal{S}} [\mathbf{R}_\gamma^t]_{s,\,l} ,
\end{equation}
{where $\overline{R}_{\gamma}^{t,\,s}$ denotes mean transfer rate for attack $\gamma$ and task $t$, when transferring adversarial examples from surrogate $s$ to all other target models $l \in\mathcal{T}\backslash\mathcal{S}$.} \textbf{For example, a higher value of $\overline{R}_{\gamma}^{t,\,s}=80\%$ would imply on an average, $80\%$ of adversarial examples that succeed on the surrogate model also succeed when transferred to the target (merged) models and MM could not defend against transfer attack. Similarly, a lower value of $\overline{R}_{\gamma}^{t,\,s}=20\%$ would mean MM successfully defended {majority} of the transfer attack}.
{Due to page limitations, we give the result of $\{\mathbf{R}_\gamma^t\}_{t=1}^{T}$ for the strongest performing attack method $\gamma=\text{NI-FGSM}$ in \tablename~\ref{tab:tr_matrix_nifgsm}, and defer the results of $\{\mathbf{R}_\gamma^t\}_{t=1}^{T},\, \forall \,\gamma\in \Gamma\backslash\{\text{NI-FGSM}\}$ to the Supplementary Material}.

{From \tablename~\ref{tab:tr_matrix_nifgsm}, we observe that when the surrogate is $f_{\boldsymbol{\theta}_0}$, the average $\overline{R}_{\gamma}^{t,\,s}$ is $66.52\%$, which is lower when compared to an average $\overline{R}_{\gamma}^{t,\,s}$ of $95.74\%$, when the surrogate is $f_{\boldsymbol{\theta}_t}$. This is expected, since fine-tuned models are more accurate and therefore serve as a better surrogate for the attack relative to pretrained models. Since the overall average $\overline{R}_{\gamma}^{t,\,s}$ is $ 81.13\%$, we claim the following:}

\begin{AIbox}{Takeaway}
\textbf{Model Merging Cannot Defend Against Transfer Attacks.} \textit{Therefore, an adversary can launch a transfer attack on an MLaaS employing a merged model with over {$\mathit{80\%}$ transfer rate on average}.}
\end{AIbox}

Next, we define the setup and process under which we statistically evaluate different hypotheses related to how transferability of adversarial examples is influenced by MM. Specifically, we evaluate three hypotheses: (1) Stronger MM methods increases transferability~($\mathcal{H}_1$) in \S\ref{subsection:hypothesis1}; (2) Mitigating representation bias via RS influences transferability~($\mathcal{H}_2$) in \S\ref{subsection:hypothesis2}; and (3) WA is the most vulnerable to transfer attacks~($\mathcal{H}_3$)  in \S\ref{subsection:hypothesis3}. In each of our hypotheses, we are interested in changes/differences in ASR as we go from one condition to another, keeping other conditions the same. For example, in $\mathcal{H}_1$, we speculate that the ASR change is positive as we go from a weaker MM method, such as TA, to AM, a relatively stronger MM method, keeping other conditions such as task $t$, surrogate model $s$ and attack method $\gamma$ same. Similarly, in $\mathcal{H}_2$, we speculate that ASR change is positive/negative as we go from (say) WA to WA+RS, keeping other the conditions same. We identify this as a related design experiment~\cite{salkind2010encyclopedia}, for which the canonical statistical tests are Student's paired $t$-test~\cite{student1908probable} and Wilcoxon signed rank test~\cite{wilcoxon1992individual}, depending on whether the sample is normal or not. Using this insight, we design a three-step statistical process to evaluate each of our hypothesis, described and illustrated in \appendixname~\ref{section:statistical_process_figure} and \figurename~\ref{fig:general_statistical_process}. 

In what follows, {$\mathbf{A}_\gamma^t \in [0, 1]^{|\mathcal{S}| \times |\mathcal{T}|}$ denotes the ASR matrix, for each task $t \in [T]$, and attack method $\gamma \in \Gamma$\footnote{ {Due to page limitations, we defer $\{\mathbf{A}_\gamma^t\}_{t=1}^{T},\, \forall \,\gamma\in \Gamma$ to the Supplementary Material.}}. Each entry $[\mathbf{A}_\gamma^t]_{s,\,l}$ denotes the ASR obtained by generating adversarial examples on surrogate model $s \in \mathcal{S}= \{f_{\boldsymbol{\theta}_0}, f_{\boldsymbol{\theta}_t}\}$ and evaluating them on target model $l \in \mathcal{T}= \mathcal{S} \cup \{f_{\boldsymbol{\theta}_{\mathit{mtl}}^m} \mid m \in \mathcal{M}\}$.} $\Delta$ denotes the difference in ASR between two conditions, $\boldsymbol{\Delta}$ denotes the tensor collecting these ASR differences, and $\mathring{\boldsymbol{\Delta}}$ denotes a set having entries of $\boldsymbol{\Delta}$ as its elements. Formally,
$\mathring{\boldsymbol{\Delta}} := \left\{ \Delta_{i_1, i_2, \dots, i_q} \;\middle|\; (i_1, i_2, \dots, i_q) \in \mathcal{I} \right\}
$ where $q$ is the \emph{rank} of the tensor (i.e., the number of dimensions), $\mathcal{I} = [n_1] \times \cdots \times [n_q]$ is the \emph{index space}, $\Delta_{i_1, \dots, i_q}$ denotes the \emph{scalar entry} $(i_1, \dots, i_q)$ of the tensor $\boldsymbol{\Delta}$. 

\subsection{Stronger Merging Increases Transferability}
\label{subsection:hypothesis1}

\begin{tcolorbox}
\textbf{Hypothesis $\boldsymbol{\mathcal{H}_1}$.} \textit{Employing a stronger MM method results in a higher transfer ASR.}
\end{tcolorbox}
 In particular, we test whether ASR increases when moving from TA or TM to AM. We do not compare TA and TM since we find no significant difference in their performance~(cf. \tablename~\ref{tab:performance_vitbase32}). 
\par
\textbf{Setup.} 
To test ${\mathcal{H}}_1$, we extract paired differences in ASR between AM and two baseline merging methods (TA and TM), both with and without RS. Specifically, for each $\gamma \in \Gamma$, task $t \in [T]$ and surrogate $s \in \mathcal{S} = \{f_{\boldsymbol{\theta}_0}, f_{\boldsymbol{\theta}_t}\}$, we compute: $\Delta^{(s, t,\gamma)}_{\text{AM, TA}} = [\mathbf{A}_{\gamma}^{t}]_{s,\, f_{\boldsymbol{\theta}_{\mathit{mtl}}^{\text{AM}}}} - [\mathbf{A}_{\gamma}^{t}]_{s, \,f_{\boldsymbol{\theta}_{\mathit{mtl}}^{\text{TA}}}},$ $\Delta^{(s, t,\gamma)}_{\text{AM, TM}} = [\mathbf{A}_{\gamma}^{t}]_{s,\,f_{\boldsymbol{\theta}_{\mathit{mtl}}^{\text{AM}}}} - [\mathbf{A}_{\gamma}^{t}]_{s,\,f_{\boldsymbol{\theta}_{\mathit{mtl}}^{\text{TM}}}},$
and collect them into tensors: $\boldsymbol{\Delta}_{\text{AM, TA}} \in [0, 1]^{|\mathcal{S}|\times T\times|\Gamma|},$ $\boldsymbol{\Delta}_{\text{AM, TM}} \in [0, 1]^{|\mathcal{S}|\times T\times|\Gamma|}.$ Similarly, for its RS-variants, we compute: $\Delta^{(s, t,\gamma)}_{\text{AM+RS, TA+RS}} = [\mathbf{A}_{\gamma}^{t}]_{s,\,f_{\boldsymbol{\theta}_{\mathit{mtl}}^{\text{AM+RS}}}} - [\mathbf{A}_{\gamma}^{t}]_{s,\,f_{\boldsymbol{\theta}_{\mathit{mtl}}^{\text{TA+RS}}}},$ $\Delta^{(s, t,\gamma)}_{\text{AM+RS, TM+RS}} = [\mathbf{A}_{\gamma}^{t}]_{s,\,f_{\boldsymbol{\theta}_{\mathit{mtl}}^{\text{AM+RS}}}} - [\mathbf{A}_{\gamma}^{t}]_{s,\,f_{\boldsymbol{\theta}_{\mathit{mtl}}^{\text{TM+RS}}}}.$
and collect them into tensors: $\boldsymbol{\Delta}^{\gamma}_{\text{AM+RS, TA+RS}} \in [0, 1]^{|\mathcal{S}| \times T} \; \forall \; \gamma \in \Gamma,$ $\boldsymbol{\Delta}_{\text{AM+RS, TA+RS}} \in [0, 1]^{|\mathcal{S}| \times T\times|\Gamma|},$ $\boldsymbol{\Delta}^{\gamma}_{\text{AM+RS, TM+RS}} \in [0, 1]^{|\mathcal{S}| \times T} \; \forall \; \gamma \in \Gamma,$ $\boldsymbol{\Delta}_{\text{AM+RS, TM+RS}} \in [0, 1]^{|\mathcal{S}| \times T\times|\Gamma|}.$


\textbf{Statistical Testing Results.} We perform statistical tests for each of the four difference sample sets ($\mathring{\boldsymbol{\Delta}}_{\text{AM, TA}}, \mathring{\boldsymbol{\Delta}}_{\text{AM, TM}}, \mathring{\boldsymbol{\Delta}}_{\text{AM+RS, TA+RS}}, $ $ \mathring{\boldsymbol{\Delta}}_{\text{AM+RS, TM+RS}}$) along with attack-specific difference sample sets ($\mathring{\boldsymbol{\Delta}}^\gamma_{\text{AM, TA}}, \mathring{\boldsymbol{\Delta}}^\gamma_{\text{AM, TM}}, $ \\ $\mathring{\boldsymbol{\Delta}}^\gamma_{\text{AM+RS, TA+RS}}, $ $\mathring{\boldsymbol{\Delta}}^\gamma_{\text{AM+RS, TM+RS}} \;\forall \;\gamma\;\in \Gamma$), while applying BH correction separately to the two groups: with and without RS. For brevity, we defer the results for attack-specific differences to the Supplementary Material. \tablename~\ref{tab:simple_models} reports the results of the statistical tests performed on two main groups. In \tablename~\ref{tab:simple_models}, we see a statistically significant difference for all of the sample sets we perform the test on. Moreover, in both the groups (with and without RS), we find the effect size for TA to be large, while the effect size for TM to be small. These observations suggest that AM is a significantly more vulnerable method than TA, while being only slightly more vulnerable than TM. 

\begin{table}[htbp]
\centering
\caption{Statistical test results for $\mathcal{H}_{1}$: stronger merging increases transferability.}
\resizebox{\linewidth}{!}{%
\begin{tabular}{lcccccc}
\toprule
\textbf{Sample} & \makecell[c]{\textbf{Shapiro–Wilk}\\\textbf{$p$-value}} & \makecell[c]{\textbf{Normal?}} & \makecell[c]{\textbf{Test}\\\textbf{Statistic}} & \makecell[c]{\textbf{One-tailed}\\\textbf{$p$-value}} & \makecell[c]{\textbf{BH-corrected}\\\textbf{$p$-value}} & \makecell[c]{\textbf{Effect}\\\textbf{Size}}\\
\midrule
\multicolumn{7}{c}{\textit{Without RS}} \\
\midrule
\rowcolor{LightGreen}
$\mathring{\boldsymbol{\Delta}}_{\text{AM, TA}}$ & $1.22 \times 10^{-2}$ & No & $W=2338.0$ & $7.22 \times 10^{-11}$ & $1.44 \times 10^{-10}$ & $r=0.766$\\
\rowcolor{LightGreen}
$\mathring{\boldsymbol{\Delta}}_{\text{AM, TM}}$ & $1.09 \times 10^{-6}$ & No & $W=1597.5$ & $1.89 \times 10^{-2}$ & $1.89 \times 10^{-2}$ & $r=0.248$\\
\midrule
\multicolumn{7}{c}{\textit{With RS}} \\
\midrule
\rowcolor{LightGreen}
$\mathring{\boldsymbol{\Delta}}_{\text{AM+RS, TA+RS}}$ & $1.90 \times 10^{-3}$ & No & $W=2410.0$ & $4.17 \times 10^{-12}$ & $8.35 \times 10^{-12}$ & $r=0.817$ \\
\rowcolor{LightGreen}
$\mathring{\boldsymbol{\Delta}}_{\text{AM+RS, TM+RS}}$ & $2.60 \times 10^{-9}$ & No & $W=1635.0$ & $1.08 \times 10^{-2}$ & $1.08 \times 10^{-2}$ & $r=0.275$ \\
\bottomrule
\end{tabular}%
}
\label{tab:simple_models}
\label{tab:surgery_models}
\end{table}

\begin{AIbox}{Takeaway}
    \textit{Employing a stronger MM method such as AM, in expectation of performance gains, would inadvertently also increase the threat of transfer attack to a large extent.}  \textit{Specifically, we observe statistical significance for all tests, with large effect size when comparing TA to AM, while only small effect size when comparing TM to AM. Therefore, AM is significantly more vulnerable than TA, while only being marginally more vulnerable than TM.}
\end{AIbox}
We further discuss the reason for this finding in \S\ref{subsection:hyp1_hyp2_discussion}.


\subsection{Mitigating Representation Bias Influences Transferability}
\label{subsection:hypothesis2}

\begin{table}[b]
\centering
\caption{Statistical test results for $\mathcal{H}_{2a}$: transferability decreases upon bias mitigation when the surrogate is a PTM.}
\resizebox{\linewidth}{!}{
\renewcommand{\arraystretch}{1.15}
\begin{tabular}{lcccccc}
\toprule
\textbf{Sample} & \makecell[c]{\textbf{Shapiro–Wilk}\\\textbf{$p$-value}} & \makecell[c]{\textbf{Normal?}} & \makecell[c]{\textbf{Test}\\\textbf{Statistic}} & \makecell[c]{\textbf{One-tailed}\\\textbf{$p$-value}} & \makecell[c]{\textbf{BH-corrected}\\\textbf{$p$-value}} & \makecell[c]{\textbf{Effect}\\\textbf{Size}} \\
\midrule
\rowcolor{LightGreen}
$\mathring{\boldsymbol{\Delta}}^{\text{FGSM}}_{\text{PTM}}$    & $4.64 \times 10^{-1}$ & Yes & $t_s = 5.55$   & $3.46 \times 10^{-6}$ & $4.33 \times 10^{-6}$ &$d=1.484$\\
\rowcolor{LightGreen}
$\mathring{\boldsymbol{\Delta}}^{\text{I-FGSM}}_{\text{PTM}}$  & $1.08 \times 10^{-1}$ & Yes & $t_s = 6.43$   & $3.43 \times 10^{-7}$ & $5.72 \times 10^{-7}$  & $d=1.718$\\
\rowcolor{LightGreen}
$\mathring{\boldsymbol{\Delta}}^{\text{PGD}}_{\text{PTM}}$ & $1.20 \times 10^{-2}$ & No & $W = 405.0$ & $7.45\times10^{-9}$ & $4.36\times 10^{-5}$ & $r=0.869$\\
\rowcolor{LightGreen}
$\mathring{\boldsymbol{\Delta}}^{\text{NI-FGSM}}_{\text{PTM}}$ & $4.19 \times 10^{-2}$ & No  & $W = 32.5$   & $1.19 \times 10^{-5}$ & $1.19 \times 10^{-5}$ & $r=0.734$\\
\rowcolor{LightGreen}
$\mathring{\boldsymbol{\Delta}}^{\text{TI-FGSM}}_{\text{PTM}}$ & $9.71 \times 10^{-3}$ & No  & $W = 11.0$   & $2.05 \times 10^{-7}$ & $5.13 \times 10^{-7}$ & $r=0.826$\\
\midrule
\rowcolor{LightGreen}
$\mathring{\boldsymbol{\Delta}}^{\text{all}}_{\text{PTM}}$ & $5.80 \times 10^{-4}$ & No  & $W = 295.5$  & $4.12 \times 10^{-17}$ & $2.06 \times 10^{-16}$ & $r=0.805$ \\
\bottomrule
\end{tabular}
}
\label{tab:hyp2a_results}
\end{table}

We divide hypothesis $\mathcal{H}_2$ into two sub-hypothesis:
\begin{tcolorbox}
\begin{enumerate}
    \item \textbf{Hypothesis $\boldsymbol{\mathcal{H}_{2a}}$.} \textit{When using pretrained model as surrogate, reducing representation bias via RS leads to a decrease in transfer ASR.}
    \item \textbf{Hypothesis $\boldsymbol{\mathcal{H}_{2b}}$.} \textit{When using fine-tuned model as surrogate, reducing representation bias via RS leads to an increase in transfer ASR.}
\end{enumerate}
\end{tcolorbox}

\textbf{Setup.} To test ${\mathcal{H}}_{2a}$, for each merging method $m_{\text{base}} \in \mathcal{M}_{\text{base}} = \{\text{WA}, \text{TA}, \text{TM}, $ $ \text{AM}\}$, we consider its RS-augmented counterpart $m_{\text{RS}} = m_{\text{base}} + \text{RS}$. For each task $t \in [T]$ and each attack method $\gamma \in \Gamma$, we compute the difference in ASR between the base model and its corresponding RS variant, under a fixed surrogate model. Specifically, when the surrogate model is a PTM $f_{\boldsymbol{\theta}_0}$, we define the difference: $\Delta^{(t, \gamma)}_{m_{\text{base}}} = [\mathbf{A}_{\gamma}^{t}]_{f_{\boldsymbol{\theta}_0},\,f_{\boldsymbol{\theta}_{\mathit{mtl}}^{m_{\text{base}}}}} -  [\mathbf{A}_{\gamma}^{t}]_{f_{\boldsymbol{\theta}_0},\,f_{\boldsymbol{\theta}_{\mathit{mtl}}^{m_{\text{RS}}}}},$ for each $m_{\text{base}} \in \mathcal{M}_{\text{base}}$. This results in $|\mathcal{M}_{\text{base}}|$ ASR differences per task, per attack method. We collect these differences across $T$ tasks to obtain the difference tensor: $\boldsymbol{\Delta}^{\gamma}_{\text{PTM}} \in [0,1]^{|\mathcal{M}_{\text{base}}|\times T}, \quad \forall \gamma \in \Gamma.$
We also collect all $|\Gamma|$ difference tensors ($\{\boldsymbol{\Delta}^{\gamma}_{\text{PTM}}\,|\, \gamma\in \Gamma\}$) into a single tensor to form: $\boldsymbol{\Delta}_{\text{PTM}}^{\text{all}} \in [0,1]^{|\mathcal{M}_{\text{base}}|\times T \times |\Gamma|}.$
To test hypothesis ${\mathcal{H}}_{2a}$, the above procedure is repeated for the case where the surrogate model is a fine-tuned model $f_{\boldsymbol{\theta}_t}$, except now, we define the difference as: $\Delta^{(t, \gamma)}_{m_{\text{base}}} =  [\mathbf{A}_{\gamma}^{t}]_{f_{\boldsymbol{\theta}_t},\,f_{\boldsymbol{\theta}_{\mathit{mtl}}^{m_{\text{RS}}}}} - [\mathbf{A}_{\gamma}^{t}]_{f_{\boldsymbol{\theta}_t},\,f_{\boldsymbol{\theta}_{\mathit{mtl}}^{m_{\text{base}}}}},$ resulting in analogous difference tensors: $\boldsymbol{\Delta}^{\gamma}_{\text{FT}} \in [0, 1]^{|\mathcal{M}_{\text{base}}|\times T}, \quad \boldsymbol{\Delta}_{\text{FT}}^{\text{all}} \in [0,1]^{|\mathcal{M}_{\text{base}}|\times T \times |\Gamma|}.$

\textbf{Statistical Testing Results.} We perform statistical tests on each of $2(|\Gamma|+1)$ difference sample sets (viz., $\{\mathring{\boldsymbol{\Delta}}^{\gamma}_{\text{PTM}}\,|\, \gamma\in \Gamma\}$, $\mathring{\boldsymbol{\Delta}}_{\text{PTM}}^{\text{all}}$, $\{\mathring{\boldsymbol{\Delta}}^{\gamma}_{\text{FT}}\,|\, \gamma\in \Gamma\}$, $\mathring{\boldsymbol{\Delta}}_{\text{FT}}^{\text{all}}$). \tablename~\ref{tab:hyp2a_results} and \tablename~\ref{tab:hyp2b_results} report the results for $\mathcal{H}_{2a}$ and $\mathcal{H}_{2b}$, respectively. From \tablename~\ref{tab:hyp2a_results}, we see a statistically significant difference for all of samples we perform the statistical test on. Moreover, we see a large effect size for all the tests. This suggests mitigating representation bias decreases vulnerability to transfer attacks to a large extent, if the surrogate model is a PTM. 
 Therefore, from \adversary's perspective which only has access to the commonly available pre-trained checkpoint of the model $f_{\boldsymbol{\theta}_0}$, RS lowers the chance of launching a successful transfer attack and is consequently undesirable. Likewise, from \owner's perspective, RS not only leads to significant accuracy gains, but also leads to an unintentional defense against weak ($f_{\boldsymbol{\theta}_0}$-equipped) attacks.
However, the results from statistical test performed for $\mathcal{H}_{2b}$ reveal a different narrative. 
In \tablename~\ref{tab:hyp2b_results}, we observe the test to be statistically significant for $\mathring{\boldsymbol{\Delta}}^{\gamma}_\text{FT} \,\forall\,\gamma \in \{\text{I-FGSM, PGD, NI-FGSM}\}$ and borderline significant for $\mathring{\boldsymbol{\Delta}}^{\text{TI-FGSM}}_\text{FT}$. Moreover, we find the effect size to be large for $\mathring{\boldsymbol{\Delta}}^{\gamma}_\text{FT} \,\forall\,\gamma \in \{\text{PGD, NI-FGSM}\}$, moderate for $\mathring{\boldsymbol{\Delta}}^{\text{I-FGSM}}_\text{FT}$ and small for $\mathring{\boldsymbol{\Delta}}^{\text{TI-FGSM}}_\text{FT}$. This suggests that mitigating bias using RS inadvertently raises the threat of transfer attack significantly when surrogate is a fine-tuned model $f_{\boldsymbol{\theta}_t}$, evident from the sample distributions. This is desirable to a stronger \adversary that employs  $f_{\boldsymbol{\theta}_t}$ as the surrogate, while being undesirable to \owner. Finally, we do not observe statistical significance for $\mathring{\boldsymbol{\Delta}}^{\text{FGSM}}_\text{FT}$. In fact, we find a moderate effect in the opposite direction, upon inspecting the effect size $r=-0.421$ in \tablename~\ref{tab:hyp2b_results} and distribution of $\mathring{\boldsymbol{\Delta}}^{\text{FGSM}}_\text{FT}$. This means an \owner using RS would be defended against an \adversary employing FGSM as her attack. This might seem contradictory to $\mathcal{H}_{2b}$ at first. However, this is expected, since we previously established from $\mathcal{H}_{2a}$ that using RS helps defend against weak attacks. FGSM is a single-step attack, and the weakest attack when compared against other iterative attacks in our work~(cf. Supplementary Material). The small effect size for TI-FGSM can similarly be explained by observing lower ASR relative to other methods~(cf. Supplementary Material). 
To further validate this, we isolate the effect of the statistically significant results, and collect $\boldsymbol{\Delta}^{\gamma}_\text{FT} \,\forall\,\gamma \in \Gamma \backslash\{\text{FGSM, TI-FGSM}\}$ into tensor
\begin{table}[h]
	\centering
	\caption{Statistical test results for $\mathcal{H}_{2b}$: transferability increases upon bias mitigation when the surrogate is a fine-tuned model. $p$-values $>$ $\alpha = 0.05$ are underlined (red).}
	\resizebox{\linewidth}{!}{
		\renewcommand{\arraystretch}{1.4}
		\begin{tabular}{lcccccc}
			\toprule
			\textbf{Sample} & \makecell[c]{\textbf{Shapiro–Wilk}\\\textbf{$p$-value}} & \makecell[c]{\textbf{Normal?}} & \makecell[c]{\textbf{Test}\\\textbf{Statistic}} & \makecell[c]{\textbf{One-tailed}\\\textbf{$p$-value}} & \makecell[c]{\textbf{BH-corrected}\\\textbf{$p$-value}} & \makecell[c]{\textbf{Effect}\\\textbf{Size}}\\
			\midrule
			\rowcolor{LightBlue}
			$\mathring{\boldsymbol{\Delta}}^{\text{FGSM}}_{\text{FT}}$    & $5.20 \times 10^{-4}$ & No  & $W = 105.0$  & \red{\underline{$9.88 \times 10^{-1}$}} & \red{\underline{$9.88 \times 10^{-1}$}} & $r=-0.421$\\
			\rowcolor{LightGreen}
			$\mathring{\boldsymbol{\Delta}}^{\text{I-FGSM}}_{\text{FT}}$  & $2.03 \times 10^{-1}$ & Yes & $t_s = 2.46$   & $1.04 \times 10^{-2}$ & $1.45 \times 10^{-2}$ & $d=0.656$\\
			\rowcolor{LightGreen}
			$\mathring{\boldsymbol{\Delta}}^{\text{PGD}}_{\text{FT}}$ & $1.23 \times 10^{-2}$ & No & $W = 375.0$ & $8.83\times10^{-6}$ & $3.09\times 10^{-5}$ & $r=0.740$\\
			\rowcolor{LightGreen}
			$\mathring{\boldsymbol{\Delta}}^{\text{NI-FGSM}}_{\text{FT}}$ & $1.00 \times 10^{-5}$ & No  & $W = 15.0$   & $1.45 \times 10^{-5}$ & $3.39\times 10^{-5}$ & $r=0.805$\\
			\rowcolor{LightBlue}
			$\boldsymbol{\Delta}^{\text{TI-FGSM}}_{\text{FT}}$ & $4.65 \times 10^{-2}$ & No  & $W = 137.0$  & \red{\underline{$6.88 \times 10^{-2}$}} & \red{\underline{$8.02 \times 10^{-2}$}} & $r=0.284$ \\
			\midrule
			\rowcolor{LightGreen}
			$\mathring{\boldsymbol{\Delta}}^{\text{all}}_{\text{FT}}$ & $2.44 \times 10^{-9}$ & No  & $W = 6772.0$ & $3.04 \times 10^{-5}$ & $5.33 \times 10^{-5}$ & $r=0.340$\\
			\rowcolor{LightGreen}
			$\mathring{\boldsymbol{\Delta}}^{\text{sig}}_{\text{FT}}$ & $6.13 \times 10^{-6}$ & No  & $W = 3067.0$ & $9.21 \times 10^{-10}$ & $6.44 \times 10^{-9}$ & $r=0.660$\\
			\bottomrule
		\end{tabular}
	}
	\label{tab:hyp2b_results}
\end{table}
$\boldsymbol{\Delta}^{\text{sig}}_\text{FT}$ and test for sample set $\mathring{\boldsymbol{\Delta}}^{\text{sig}}_{\text{FT}} = \mathring{\boldsymbol{\Delta}}^{\text{all}}_{\text{FT}}\backslash\{\mathring{\boldsymbol{\Delta}}^{\text{FGSM}}_{\text{FT}} \cup \mathring{\boldsymbol{\Delta}}^{\text{TI-FGSM}}_{\text{FT}}\}$. We find a statistically significant result for $\mathring{\boldsymbol{\Delta}}^{\text{sig}}_\text{FT}$ with large effect size as seen in \tablename~\ref{tab:hyp2b_results}.

\begin{AIbox}{Takeaway}
\textit{Mitigating representation bias can only defend against weaker attacks~(e.g., attacks employing a weaker surrogate $f_{\boldsymbol{\theta}_0}$, or employing a weaker attack method, such as FGSM). In the face of a strong attack~(e.g., attacks employing a stronger surrogate $f_{\boldsymbol{\theta}_t}$ or a strong attack method, such as PGD or NI-FGSM), the threat of transfer attack increases significantly for owner. Indeed, there is no such thing as a ``free lunch''.}
\end{AIbox}
We further discuss the reason for this finding in \S\ref{subsection:hyp1_hyp2_discussion}.\vspace{-1em}
\subsection{Weight Averaging is Most Vulnerable to Transfer Attacks}
\label{subsection:hypothesis3}
\begin{table}[b]
\centering
\caption{Statistical test results for $\mathcal{H}_{3}$: WA is the most vulnerable MM method.}
\resizebox{\linewidth}{!}{
\begin{tabular}{lcccccc}
\toprule
\textbf{Sample} & \makecell[c]{\textbf{Shapiro–Wilk}\\\textbf{$p$-value}} & \makecell[c]{\textbf{Normal?}} & \makecell[c]{\textbf{Test}\\\textbf{Statistic}} & \makecell[c]{\textbf{One-tailed}\\\textbf{$p$-value}} & \makecell[c]{\textbf{BH-corrected}\\\textbf{$p$-value}} & \makecell[c]{\textbf{Effect}\\\textbf{Size}} \\
\midrule
\multicolumn{7}{c}{\textit{Without RS: WA is more vulnerable than TA, TM, AM}} \\
\midrule
\rowcolor{LightGreen}
$\mathring{\boldsymbol{\Delta}}_{\text{WA},\, \text{TA}}$         & $4.36 \times 10^{-2}$ & No  & $W = 1563.0$ & $2.19 \times 10^{-10}$ & $6.57 \times 10^{-10}$ & $r=0.810$\\
\rowcolor{LightGreen}
$\mathring{\boldsymbol{\Delta}}_{\text{WA},\, \text{TM}}$         & $8.12 \times 10^{-3}$ & No  & $W = 1473.0$ & $1.84 \times 10^{-8}$  & $2.76 \times 10^{-8}$ & $r=0.710$  \\
\rowcolor{LightGreen}
$\mathring{\boldsymbol{\Delta}}_{\text{WA},\, \text{AM}}$         & $2.80 \times 10^{-4}$ & No  & $W = 1215.0$ & $3.35 \times 10^{-4}$  & $3.35 \times 10^{-4}$  & $r=0.444$\\
\midrule
\multicolumn{7}{c}{\textit{With RS: WA+RS is more vulnerable than TA+RS, TM+RS, AM+RS}} \\
\midrule
\rowcolor{LightGreen}
$\mathring{\boldsymbol{\Delta}}_{\text{WA+RS},\, \text{TA+RS}}$         & $6.95 \times 10^{-3}$ & No  & $W = 1557.0$  & $2.98 \times 10^{-10}$ & $8.94 \times 10^{-10}$ & $r=0.821$ \\
\rowcolor{LightGreen}
$\mathring{\boldsymbol{\Delta}}_{\text{WA+RS},\, \text{TM+RS}}$         & $2.25 \times 10^{-2}$ & No  & $W = 1478.0$  & $1.45 \times 10^{-8}$  & $2.18 \times 10^{-8}$  & $r=0.740$\\
\rowcolor{LightGreen}
$\mathring{\boldsymbol{\Delta}}_{\text{WA+RS},\, \text{AM+RS}}$         & $3.00 \times 10^{-6}$ & No  & $W = 1281.5$  & $9.00 \times 10^{-6}$  & $9.00 \times 10^{-6}$  & $r=0.585$ \\
\bottomrule
\end{tabular}
}
\label{tab:wa_simple_vs_others}
\label{tab:wa_surgery_vs_others}
\end{table}

\begin{tcolorbox}
\textbf{Hypothesis $\boldsymbol{\mathcal{H}_{3}}$.} \textit{WA yields the highest transfer ASR despite being the weakest MM method.}
\end{tcolorbox}

\textbf{Setup.} For each task $t \in [T]$, surrogate model $s \in \mathcal{S} = \{f_{\boldsymbol{\theta}_0}, f_{\boldsymbol{\theta}_t}\}$, and attack method $\gamma \in \Gamma$, we compare WA against the other base merging methods: TA, TM, and AM. We define the difference in ASR between WA and another merging method $m \in \{\text{TA}, \text{TM}, \text{AM}\} = \mathcal{M}_{\text{WA}'}$ for task $t$ and surrogate $s$ as: $\Delta_{\text{WA},\, m}^{(s, t, \gamma)} = [\mathbf{A}_{\gamma}^{t}]_{s,\,f_{\boldsymbol{\theta}_{\mathit{mtl}}^{\text{WA}}}} - [\mathbf{A}_{\gamma}^{t}]_{s,\,f_{\boldsymbol{\theta}_{\mathit{mtl}}^m}}.$ We compute $\Delta_{\text{WA}, m}^{(s, t, \gamma)}$ for both $s \in \mathcal{S}$ and for each $t \in [T]$, resulting in $|\mathcal{S}|$ values per task. Collecting these differences across $T$ tasks yields the difference tensor: $\boldsymbol{\Delta}_{\text{WA},\, m}^{\gamma} \in [0,1]^{|\mathcal{S}|\times T}, \forall m \in \mathcal{M}_{\text{WA}'}.$
Finally, we collect all difference tensors $|\Gamma|$ $\{\boldsymbol{\Delta}_{\text{WA}, m}^{\gamma} \,|\,\gamma \in \Gamma\}$ to form the tensor: $\boldsymbol{\Delta}_{\text{WA}, m} \in [0,1]^{|\mathcal{S}|\times T \times |\Gamma|}.$
We repeat this process independently for RS-augmented variants. For the RS variants, the difference is defined as: $\Delta_{\text{WA+RS},\, m+\text{RS}}^{(s, t, \gamma)} = [\mathbf{A}_{\gamma}^{t}]_{s,\,f_{\boldsymbol{\theta}_{\mathit{mtl}}^{\text{WA+RS}}}} - [\mathbf{A}_{\gamma}^{t}]_{s,\,f_{\boldsymbol{\theta}_{\mathit{mtl}}^{m+\text{RS}}}},$ with corresponding difference tensors: $\boldsymbol{\Delta}_{\text{WA+RS},\, m+\text{RS}}^{\gamma} \in [0,1]^{|\mathcal{S}|\times T}, \forall m \in \mathcal{M}_{\text{WA}'},$ $\boldsymbol{\Delta}_{\text{WA+RS}, \,m+\text{RS}} \in [0,1]^{|\mathcal{S}|\times T \times |\Gamma|}, \forall m \in \mathcal{M}_{\text{WA}'}.$

\textbf{Statistical Testing Result.} We perform statistical tests for $\mathring{\boldsymbol{\Delta}}^{\gamma}_{\text{WA},\, m}$,
 \\
 $\mathring{\boldsymbol{\Delta}}^\gamma_{\text{WA+RS},\, m+{\text{RS}}}$, $\mathring{\boldsymbol{\Delta}}_{\text{WA},\, m}$, $\mathring{\boldsymbol{\Delta}}_{\text{WA+RS},\, m+{\text{RS}}}$ $\forall \,m \in \mathcal{M}_{\text{WA}'}\;\forall \; \gamma \in \Gamma$, and apply BH correction to each of the groups~(with and without RS) separately. For brevity, we defer results of attack-specific samples to the Supplementary Material. \tablename~\ref{tab:wa_simple_vs_others} shows the results for $\mathring{\boldsymbol{\Delta}}_{\text{WA},\, m}$ and $\mathring{\boldsymbol{\Delta}}_{\text{WA+RS},\, m+{\text{RS}}}$ $\forall \,m \in \mathcal{M}_{\text{WA}'}$. We find statistically significant results for all sample sets we perform the test on. Moreover, we find a large effect size for all samples except $\mathring{\boldsymbol{\Delta}}_{\text{WA},\, \text{AM}}$ where the effect size is medium, indicating WA increases the threat of transfer attack significantly.


\begin{AIbox}{Takeaway}
\textit{Despite being the weakest MM method, WA increases the threat of transfer attack the most. Consequently, an owner suffers not only from poor multi-task model performance when employing WA, but also increases her threat to transfer attacks the most.}
\end{AIbox}
We further discuss the reason for this finding in \S\ref{subsection:hyp3_discussion}. The results for all hypotheses, viz., $\mathcal{H}_1$, $\mathcal{H}_2$ and $\mathcal{H}_3$ are summarized in \figurename~\ref{fig:radar_summary}.

\begin{figure}[!htbp]
    \centering
    \begin{subfigure}[t]{0.48\linewidth}
        \centering
        \resizebox{\linewidth}{!}{
\begin{tikzpicture}[scale=1]
\def\R{3.5}  
\def\maxval{12}

\pgfmathsetmacro{\aA}{90}
\pgfmathsetmacro{\aB}{90-72}
\pgfmathsetmacro{\aC}{90-144}
\pgfmathsetmacro{\aD}{90-216}
\pgfmathsetmacro{\aE}{90-288}

\foreach \lev in {3,6,9,12} {
  \pgfmathsetmacro{\rr}{\lev/\maxval*\R}
  \draw[gray!35, thin]
    (\aA:\rr) -- (\aB:\rr) -- (\aC:\rr) -- (\aD:\rr) -- (\aE:\rr) -- cycle;
}

\foreach \ang in {\aA,\aB,\aC,\aD,\aE} {
  \draw[gray!50, thin] (0,0) -- (\ang:\R);
}

\foreach \lev in {3,6,9,12} {
  \pgfmathsetmacro{\rr}{\lev/\maxval*\R}
  \node[font=\LARGE\bfseries, fill=white, inner sep=0.5pt, text=gray!60] at (90:\rr+0.15) {\lev};
}

\node[font=\LARGE\bfseries, anchor=south] at (\aA:\R+0.3) {AM$-$TA\,{\LARGE($\mathcal{H}_1$)}};
\node[font=\LARGE\bfseries, anchor=west] at (\aB:\R+0.2) {AM$-$TM\,{\LARGE($\mathcal{H}_1$)}};
\node[font=\LARGE\bfseries, anchor=north west] at (\aC:\R+0.2) {WA$-$TA\,{\LARGE($\mathcal{H}_3$)}};
\node[font=\LARGE\bfseries, anchor=north east] at (\aD:\R+0.2) {WA$-$TM\,{\LARGE($\mathcal{H}_3$)}};
\node[font=\LARGE\bfseries, anchor=east] at (\aE:\R+0.2) {WA$-$AM\,{\LARGE($\mathcal{H}_3$)}};

\pgfmathsetmacro{\rAa}{5.19/\maxval*\R}
\pgfmathsetmacro{\rAb}{0.72/\maxval*\R}
\pgfmathsetmacro{\rAc}{9.36/\maxval*\R}
\pgfmathsetmacro{\rAd}{4.39/\maxval*\R}
\pgfmathsetmacro{\rAe}{1.75/\maxval*\R}

\draw[blue!70!black, thick, fill=blue!70!black, fill opacity=0.15]
  (\aA:\rAa) -- (\aB:\rAb) -- (\aC:\rAc) -- (\aD:\rAd) -- (\aE:\rAe) -- cycle;
\fill[blue!70!black] (\aA:\rAa) circle (2pt);
\fill[blue!70!black] (\aB:\rAb) circle (2pt);
\fill[blue!70!black] (\aC:\rAc) circle (2pt);
\fill[blue!70!black] (\aD:\rAd) circle (2pt);
\fill[blue!70!black] (\aE:\rAe) circle (2pt);

\pgfmathsetmacro{\rBa}{5.74/\maxval*\R}
\pgfmathsetmacro{\rBb}{0.32/\maxval*\R}
\pgfmathsetmacro{\rBc}{9.76/\maxval*\R}
\pgfmathsetmacro{\rBd}{3.89/\maxval*\R}
\pgfmathsetmacro{\rBe}{2.29/\maxval*\R}

\draw[red!70!black, thick, dashed, fill=red!70!black, fill opacity=0.08]
  (\aA:\rBa) -- (\aB:\rBb) -- (\aC:\rBc) -- (\aD:\rBd) -- (\aE:\rBe) -- cycle;
\fill[red!70!black] (\aA:\rBa) circle (2pt);
\fill[red!70!black] (\aB:\rBb) circle (2pt);
\fill[red!70!black] (\aC:\rBc) circle (2pt);
\fill[red!70!black] (\aD:\rBd) circle (2pt);
\fill[red!70!black] (\aE:\rBe) circle (2pt);

\node[anchor=north west, draw, rounded corners, fill=white, inner sep=3pt, font=\LARGE\bfseries]
  at (2.8, 3.8) {
  \begin{tabular}{@{}c@{\hspace{3pt}}l@{}}
    \tikz\draw[blue!70!black, thick] (0,0) -- (0.4,0); & w/o RS \\
    \tikz\draw[red!70!black, thick, dashed] (0,0) -- (0.4,0); & w/ RS \\
  \end{tabular}
};

\end{tikzpicture}
        }
        \caption{$\mathcal{H}_1$ and $\mathcal{H}_3$: median ASR difference (\%) for MM method comparisons.}
        \label{fig:radar_h1h3}
    \end{subfigure}
    \hfill
    \begin{subfigure}[t]{0.48\linewidth}
        \centering
        \resizebox{\linewidth}{!}{
\begin{tikzpicture}[scale=1]
\def\R{3.5}  
\def\maxval{6}

\pgfmathsetmacro{\aA}{90}
\pgfmathsetmacro{\aB}{90-72}
\pgfmathsetmacro{\aC}{90-144}
\pgfmathsetmacro{\aD}{90-216}
\pgfmathsetmacro{\aE}{90-288}

\foreach \lev in {2,4,6} {
  \pgfmathsetmacro{\rr}{\lev/\maxval*\R}
  \draw[gray!35, thin]
    (\aA:\rr) -- (\aB:\rr) -- (\aC:\rr) -- (\aD:\rr) -- (\aE:\rr) -- cycle;
}

\foreach \ang in {\aA,\aB,\aC,\aD,\aE} {
  \draw[gray!50, thin] (0,0) -- (\ang:\R);
}

\foreach \lev in {2,4,6} {
  \pgfmathsetmacro{\rr}{\lev/\maxval*\R}
  \node[font=\LARGE\bfseries, fill=white, inner sep=0.5pt, text=gray!60] at (90:\rr+0.15) {\lev};
}

\node[font=\LARGE\bfseries, anchor=south] at (\aA:\R+0.3) {FGSM};
\node[font=\LARGE\bfseries, anchor=west] at (\aB:\R+0.2) {I-FGSM};
\node[font=\LARGE\bfseries, anchor=north west] at (\aC:\R+0.2) {PGD};
\node[font=\LARGE\bfseries, anchor=north east] at (\aD:\R+0.2) {NI-FGSM};
\node[font=\LARGE\bfseries, anchor=east] at (\aE:\R+0.2) {TI-FGSM};

\pgfmathsetmacro{\rAa}{4.71/\maxval*\R}
\pgfmathsetmacro{\rAb}{4.44/\maxval*\R}
\pgfmathsetmacro{\rAc}{4.41/\maxval*\R}
\pgfmathsetmacro{\rAd}{2.50/\maxval*\R}
\pgfmathsetmacro{\rAe}{5.11/\maxval*\R}

\draw[green!50!black, thick, fill=green!50!black, fill opacity=0.15]
  (\aA:\rAa) -- (\aB:\rAb) -- (\aC:\rAc) -- (\aD:\rAd) -- (\aE:\rAe) -- cycle;
\fill[green!50!black] (\aA:\rAa) circle (2pt);
\fill[green!50!black] (\aB:\rAb) circle (2pt);
\fill[green!50!black] (\aC:\rAc) circle (2pt);
\fill[green!50!black] (\aD:\rAd) circle (2pt);
\fill[green!50!black] (\aE:\rAe) circle (2pt);

\pgfmathsetmacro{\rBa}{0.01/\maxval*\R}
\pgfmathsetmacro{\rBb}{1.23/\maxval*\R}
\pgfmathsetmacro{\rBc}{1.55/\maxval*\R}
\pgfmathsetmacro{\rBd}{0.99/\maxval*\R}
\pgfmathsetmacro{\rBe}{1.31/\maxval*\R}

\draw[orange!80!black, thick, dashed, fill=orange!80!black, fill opacity=0.08]
  (\aA:\rBa) -- (\aB:\rBb) -- (\aC:\rBc) -- (\aD:\rBd) -- (\aE:\rBe) -- cycle;
\fill[orange!80!black] (\aA:\rBa) circle (2pt);
\fill[orange!80!black] (\aB:\rBb) circle (2pt);
\fill[orange!80!black] (\aC:\rBc) circle (2pt);
\fill[orange!80!black] (\aD:\rBd) circle (2pt);
\fill[orange!80!black] (\aE:\rBe) circle (2pt);

\node[anchor=north west, draw, rounded corners, fill=white, inner sep=3pt, font=\LARGE\bfseries]
  at (2.5, 3.8) {
  \begin{tabular}{@{}c@{\hspace{3pt}}l@{}}
    \tikz\draw[green!50!black, thick] (0,0) -- (0.4,0); & $\mathcal{H}_{2a}$: PTM \\
    \tikz\draw[orange!80!black, thick, dashed] (0,0) -- (0.4,0); & $\mathcal{H}_{2b}$: FT \\
  \end{tabular}
};

\end{tikzpicture}
        }
        \caption{$\mathcal{H}_2$: median ASR difference (\%) by attack method upon bias mitigation.}
        \label{fig:radar_h2}
    \end{subfigure}
    \caption{Radar plots summarizing median ASR differences across all hypotheses. (a)~$\mathcal{H}_1$/$\mathcal{H}_3$: WA$-$TA shows the largest difference, confirming WA is most vulnerable; AM$-$TA confirms stronger merging increases transferability. (b)~$\mathcal{H}_2$: RS defends against PTM-based attacks (green) but increases vulnerability to FT-based attacks (orange).}
    \label{fig:radar_summary}
\end{figure}

\section{Related Works}
\label{section:related_work}
Early works~\cite{frankle2020linear,izmailov2018averaging,neyshabur2020being,ilharco2022patching} demonstrate that when two neural networks share part of their optimization trajectory, their weights can often be linearly interpolated without degrading performance---a phenomenon known as \textit{Linear Mode Connectivity}~(LMC). Subsequent studies extend this to feature space, introducing \textit{Layerwise Linear Feature Connectivity}~(LLFC), where the interpolated model features are proportional to those of its endpoints~\cite{llfc}. More recently, Zhou et al.~\cite{ctl} show that LLFC holds in the pretraining-finetuning paradigm, and identify a stronger notion of linearity than LLFC, termed \textit{Cross Task Linearity}~(CTL), which asserts the features in the weight-interpolated model are approximately equal
to the linear interpolation of features in the two finetuned
models at each layer. In doing so, they provide explanation for why MM methods such as WA and TA work. These findings offer a new perspective on MM, which has been employed for enhancing generalization {and out-of-distribution robustness}~\cite{yang2024model,reg_mean}. They are also used in constructing multi-task models~\cite{task_arithmetic,ties_merging,ada_merging,representation_surgery}, and supporting domains like parameter-efficient tuning~\cite{zhang2023composing}, reinforcement learning~\cite{rame2024rewarded}, and diffusion models~\cite{nair2024maxfusion}.

Despite its broad applicability, the security implications of MM has largely remained unexplored, with only few works recently examining it. {Out of these works, most either show how MM enhances robustness against various attacks, or disrupt the MM pipeline to launch adversarial attacks.} 
{For instance, Wang et al.~\cite{wang2024rethinking} show that weight averaged models reduce absolute weight values and parameter variance, implying WA could decrease the upper bounds on overall Lipschitz constant of a network, and consequently increase robustness to adversarial examples. Zhou et al.~\cite{zhou2025mixup} propose Mixup Model Merge, an MM method inspired from mixup data augmentation that enhances out-of-distribution and adversarial robustness along with performance. Farn et al.~\cite{farn2024safeguard} show that merging weights of pretrained and fine-tuned models preserves safety and robustness while also enhancing performance.}  Arora et al.~\cite{free_lunch} are first to observe MM mitigates backdoors. Specifically, by merging a poisoned model with a clean counterpart using WA, the merged model maintains high clean accuracy while significantly reducing backdoor ASR, thereby giving a ``free lunch'' of robustness. 

Concurrently, BadMerging~\cite{zhang2024badmerging} also notes vanilla MM mitigates backdoor attacks, and designs persistent backdoors in the individual fine-tuned models that survive merging process by crafting a universal trigger. With this, BadMerging achieves up to 95\% ASR. MergeBackdoor~\cite{wang2025purity} introduces a novel supply-chain threat for MM, in which seemingly benign models when merged result in backdoored model. Concurrently, Merge Hijacking~\cite{yuan-etal-2025-merge} introduces the first backdoor attack against MM in large language models. In these works, vanilla MM pipeline is assumed to be harmless, with adversary disrupting the MM pipeline for launching an attack. In contrast, {our work is the first to highlight} vulnerabilities inherent to MM, i.e., without ever disrupting the MM pipeline, we show how MM is not just susceptible, but increases the threat to transfer attacks.

The most related work to our is by Gangwal and Sharma~\cite{gangwal2025poster}, which investigates transferability of adversarial examples in MM. However, their preliminary analysis consists of only two MM methods and three datasets, and does not consist of rigorous statistical evaluations as ours. Moreover, it alludes to MM providing ``free'' adversarial robustness, similar to Arora et al., which is the notion we challenge. Our work is, to our best knowledge, the first to systematically and extensively study the risk of transfer attacks in the context of MM methods.
{We show that MM increases the threat to transferable adversarial examples, and provide actionable insights for ML practitioners designing reliable ML systems.}
 
\section{Discussion}
\label{section:discussion}
\subsection{Stronger Merging Methods, Representation Surgery and Transferability}
\label{subsection:hyp1_hyp2_discussion}
\textbf{Explaining Higher Transferability for Stronger Merging Methods ($\boldsymbol{\mathcal{H}_1}$).}  In \S\ref{subsection:hypothesis1}, we observe {that} stronger MM methods tend to have higher transferability. Prior works have shown decision boundary similarity~\cite{tramer2017space,liu2017delving} and higher accuracy of target model~\cite{higher_test_transferability1,yang2021trs} are correlated with higher transferability. Since the goal of MM is minimizing the loss on the test set across multiple tasks~(cf. Eq.~\eqref{eq:model_merging_goal}), it explains higher transferability of stronger MM methods.

\textbf{Explaining the Effect of Representation Bias in Transferability ($\boldsymbol{\mathcal{H}_{2a}}$ and $\boldsymbol{\mathcal{H}_{2b}}$).} In \S\ref{subsection:hypothesis2}, we observe that when the surrogate is a PTM, reducing representation bias leads to lower transferability~($\mathcal{H}_{2a}$), while when the surrogate is a fine-tuned model, reducing representation bias leads to higher transferability~($\mathcal{H}_{2b}$). To explain this, we observe that minimizing the objective in Eq.~\eqref{eq:representation_surgery} pulls representations obtained from the merged model ($\widehat{\boldsymbol{{Z}}}_{\mathit{mtl}}^t$) closer to representations obtained from the individual, fine-tuned models ($\boldsymbol{Z}_{\mathit{ind}}^t$). Since during classification, the classifier head for every task is reused in the merged model, it means reducing representation bias is the same as making the merged model's decision boundary closer to the fine-tuned models. This explains $\mathcal{H}_{2b}$, since now the target (RS-augmented) model's decision boundary aligns more surrogate (fine-tuned) model than its RS-free counterpart. It also explains $\mathcal{H}_{2a}$, since it simultaneously makes the target (RS-augmented) model's decision boundary dissimilar to PTM's.



\subsection{Weight Averaging and Transferability}
\label{subsection:hyp3_discussion}
\textbf{Explaining Higher Transferability of Weight Averaging ($\boldsymbol{\mathcal{H}_{3}}$).} In \S\ref{subsection:hypothesis3}, we observe WA consistently exhibits higher transferability than all other MM methods, despite being the weakest merging method in terms of test accuracy~\cite{representation_surgery}. To understand this, we look at CTL~\cite{ctl}, which shows that for each layer $\ell$ and input $\mathbf{x}$,
\begin{equation}
\label{eq:ctl}
 f^{(\ell)}_{\boldsymbol{\theta}_{mtl}^{\text{WA}}}(\mathbf{x})
 \;=\;
 f^{(\ell)}_{\frac{1}{T}\sum_t\boldsymbol{\theta}_t}(\mathbf{x})
 \;\approx\;
 \frac{1}{T}\sum_{t=1}^T f^{(\ell)}_{\boldsymbol{\theta}_t}(\mathbf{x}).
\end{equation}
If we set $\ell=L$ (the last layer) in Eq.~\eqref{eq:ctl}, we find the prediction of WA approximates the logit ensemble of individual models. We conjecture this enables WA model's gradient to be approximately at the center of individual models' gradient directions\footnote{{We give an alternate perspective on explanation for $\mathcal{H}_3$ from model ensemble literature in \appendixname~\ref{section:ensemble_discussion}.}}. Mathematically,   
\begin{equation}
\label{eq:wa_conjecture}
{\nabla_{\mathbf{x}}\mathcal{L}\bigl(f_{\boldsymbol{\theta}_{mtl}^{\text{WA}}}(\mathbf{x}),\mathbf{y}\bigr)
}\;\approx\;
\frac{1}{T}\sum_{t=1}^T \frac{\nabla_{\mathbf{x}}\mathcal{L}\bigl(f_{\boldsymbol{\theta}_t}(\mathbf{x}),\mathbf{y}\bigr)}{|\nabla_{\mathbf{x}}\mathcal{L}\bigl(f_{\boldsymbol{\theta}_t}(\mathbf{x}),\mathbf{y}\bigr)|}.
\end{equation}
Consequently, this maximizes the average cosine similarity between surrogate and target model's gradient~(\cf \appendixname~\ref{section:proof} for proof),  and leads to high alignment between them, which is a well-known driver for transferability~\cite{higher_test_transferability1,papernot2017practical,demontis2019adversarial}. This makes WA more susceptible to adversarial transfer than other MM methods that move the target model's gradient direction away from the center with scaling coefficient $\lambda \neq 1/T$. 

We test this conjecture by checking whether a merged model's input gradient points to the empirical center of the individual gradient directions. For a sample $(\mathbf{x},\mathbf{y})$ let $g_t=\nabla_{\mathbf{x}}\mathcal{L}\bigl(f_{\boldsymbol{\theta}_t}(\mathbf{x}),\mathbf{y}\bigr),~
\hat g_t=\frac{g_t}{\|g_t\|},~
\bar g=\tfrac{1}{T}\sum_{t=1}^T \hat g_t,~
\hat{\bar g}=\bar g/\|\bar g\|.$
Then for a merged model $M$ with unit gradient $\hat g_M$, define the \emph{center} score $c$ as $\mathrm{c}(M)=\langle \hat g_M,\hat{\bar g}\rangle.$ We find the mean $c$ to be 0.0283 and 0.0087 for WA and other methods, respectively, indicating WA's gradient are positioned relatively closer towards center of individual models' unit gradient than other methods. We perform a Mann-Whitney U test~\cite{mann1947test} to confirm the statistical significance and find $p$-value to be $1.832\times10^{-6}$, indicating the difference is non-trivial. Finally, we also calculate the alignment between the gradients of surrogate and target models for each method with mean cosine similarity between them. We find the score for WA and other methods to be 0.0827 and 0.0089, respectively with a Mann-Whitney $p$-value $8.131\times10^{-47}$, indicating surrogate model's gradients do indeed align better with WA than other methods, making WA naturally more susceptible to transfer attacks than other methods.

\subsection{Targeted Attacks}
\label{subsection:targeted_attacks}
Throughout the paper, we have assumed the attacks to be untargeted. However, an adversary may have incentive to perform targeted attacks. Therefore, we perform targeted attacks using $\gamma=\text{NI-FGSM}\;\forall $  target models and datasets. {The Supplementary Material presents the result of targeted attacks performed. When considering pretrained model as surrogate, we find the average $\overline{R}_{\gamma}^{t,\,s}$ to only $0.14\%$. This indicates MM is able to successfully defend against targeted attacks performed using pretrained models as surrogate, consistent with $\mathcal{H}_{2a}$ which posits MM can defend against weak attacks. We also find $\overline{R}_{\gamma}^{t,\,s}=80.55\%$ when the surrogate is fine-tuned model, indicating MM cannot defend against targeted attacks performed using stronger fine-tuned models as the surrogate, consistent with $\mathcal{H}_{2b}$.}

\subsection{{Towards Realistic Attacks with Limited Information}}
\label{subsection:limited_information_attacks}
{So far, we have assumed {the} adversary to know owner uses a merged model, the architecture of the merged model, and having access to the corresponding pretrained/fine-tuned model. These are strong assumption, and we acknowledge our evaluations likely reflect an upper bound of vulnerabilities existing within current MM methods. Even so, we consider three realistic scenarios where an adversary attacks the merged model with limited information: (1) with a surrogate of different architecture, as knowing the architecture may not always be feasible, despite ready availability of pretrained/fine-tuned models of corresponding architecture; (2) with a fine-tuned surrogate that is independently trained, as the fine-tuned models that are used to attack may not be trained on the same data as the fine-tuned models which are used in creation of the target merged model, and (3) with no surrogate model and no knowledge of MM, using query-based black-box attacks. Since query-based attacks require almost no knowledge, they are hardest attacks to perform, and if successful, equally hard to defend against. We provide results for these three as follows{, and summarize the results in \tablename~\ref{tab:limited_info_attacks}.}}

\subsubsection{{Attack with Different Surrogate}}
{We present the result on CLIP models with ResNet-50 backbone in the Supplementary Material. Here, we observe that when the surrogate is pretrained, average $\overline{R}_{\gamma}^{t,\,s}$ is $40.56\%$. 
Concurrently, when surrogate is fine-tuned, we observe average $\overline{R}_{\gamma}^{t,\,s}$ to be $59.94\%$. 
Overall, since the average $\overline{R}_{\gamma}^{t,\,s}$ taken over all surrogates is $50.25\%$, we claim MM cannot reliably defend against transfer attacks even with a different surrogate.}

\subsubsection{Attack with Independently Trained Surrogate}
{We split the training data into two disjoint halves. From the first disjoint half, we obtained fine-tuned models that we used specifically for the creation of different merged models. Then, we used only those fine-tuned models that were obtained from the second disjoint half as surrogate to perform the attack. The results are present in the Supplementary Material. We observe that the average $\overline{R}_{\gamma}^{t,\,s}$ when the surrogate is fine-tuned is $95.10\%$, which is not significantly different from the original setting, where we observed average $\overline{R}_{\gamma}^{t,\,s}$ of $95.74\%$. Therefore, even under this limited setting, we show that MM cannot defend against transfer attacks.} 

\subsubsection{Query-based (black-box) Attacks}
 We employ the Square Attack~\cite{andriushchenko2020square} to attack the merged models. We present the results of attack performed in the Supplementary Material. Here, we find the ASR is no less than 80\% for 43/56 cases, with average ASR $=85.00\%$, indicating MM cannot reliably defend against query-based black-box attacks.
 
 \begin{table}[htbp]
 	\centering
 	\caption{{Summary of attack results under limited-information setting.}}
 	\label{tab:limited_info_attacks}
 	\setlength{\tabcolsep}{3.5pt}
 	\resizebox{\linewidth}{!}{%
 		\begin{tabular}{l l l c l}
 			\toprule
 			\makecell[c]{\textbf{Threat Setting}} & \makecell[c]{\textbf{Adversary Access}} & \makecell[c]{\textbf{Surrogate/} \\ \textbf{Method}} & \makecell[c]{\textbf{$\overline{R}_{\gamma}^{t,\,s}$/ASR}} & \makecell[c]{\textbf{Key Takeaway}} \\
 			\midrule
 			\textbf{Cross-Architecture} &
 			\makecell[l]{Different backbone\\architecture} &
 			\makecell[l]{ResNet-50 $\rightarrow$\\ViT-B/16} &
 			40.0\%--60.0\% &
 			Not reliably defended \\
 			\midrule
 			\textbf{Disjoint Data} &
 			\makecell[l]{Zero shared samples\\b/w surrogate \& target} &
 			\makecell[l]{Fine-tuned $f_{\boldsymbol{\theta}_t}$s\\on disjoint split} &
 			$>95.0\%$ &
 			Not defended \\
 			\midrule
 			\textbf{Query-Based} &
 			\makecell[l]{Zero gradient /\\architecture access} &
 			\makecell[l]{Square Attack\\($N=500$)} &
 			85.0\% &
 			Not defended \\
 			\bottomrule
 		\end{tabular}%
 	}
 \end{table}   

\subsection{Towards Mitigation and Robust Model Merging}
\label{subsection:solution_and_conjecture}
\begin{wrapfigure}{r}{0.5\linewidth}
    \centering
    \vspace{-1em}
    \includegraphics[width=\linewidth]{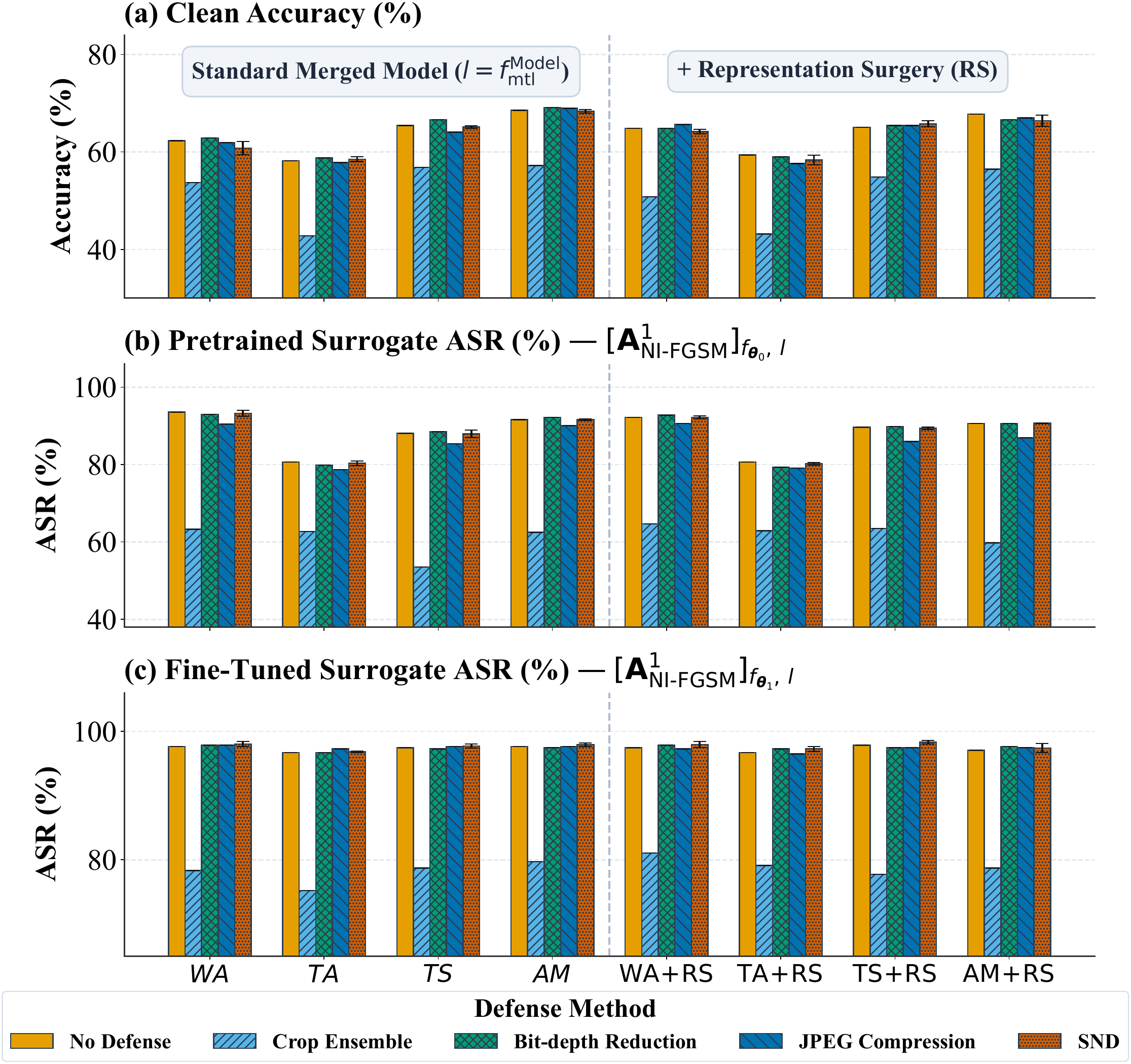}
    \caption{Result of input transformation defenses. Here, $\gamma=\text{NI-FGSM}$.}
    \label{fig:input_level_defense}
    \vspace{-1em}
\end{wrapfigure}

MM offers multi-task capability without joint training or access to all task data, making it a cost-effective alternative to multi-task learning~(MTL). However, there are no works analyzing transferability of adversarial examples~(and defense) in the context of MTL. There are few works that address white-box adversarial robustness in MTL~\cite{liu2017delving,mao2020multitask,zhang2023multi,ghamizi2022adversarial}, {along with some that show promise in demonstrating how models could be made robust to different attacks using MM~\cite{croce2023seasoning,toishi2025enhancing}. However, both are not directly applicable, as they require adversarially trained model, which cannot be obtained due to lack of training data access in MM setting.} Therefore, we test the efficacy of common input transformation defenses against adversarial examples. We test three input transformation defenses from Guo et al.~\cite{guo2018countering}: (1) Crop Ensemble, i.e., cropping and rescaling images and averaging prediction over multiple crops; (2) Bit-depth Reduction; and (3) JPEG Compression. We also test SND~\cite{byun2022effectiveness}~(adding small Gaussian noise to input), an input transformation defense against query-based attacks due to its simplicity\footnote{{We exclude randomized smoothing based defenses~\cite{levine2020randomized,cohen2019certified} since they are known to be ineffective against $\ell_\infty$ attacks~\cite{yang2020randomized,blum2020random,kumar2020curse}.}}. Figure~\ref{fig:input_level_defense} presents the results with two surrogates: pretrained model~($f_{\boldsymbol{\theta}_0}$) and model fine-tuned on $(t=1)$ Cars~($f_{\boldsymbol{\theta}_1}$) with $\gamma=\text{NI-FGSM}$ as the attack method.
Clearly, we observe no significant change in ASR for Bit-depth Reduction, JPEG Compression or SND. Therefore, these defenses prove to be ineffective. We do observe a modest drop in ASR for Crop Ensemble; however, not only is it computationally prohibitive since it requires multiple~(30) inferences for a single image, it is also accompanied by significant drop in the model utility (clean accuracy), making the defense impractical.

Inspired from the literature, we suggest the following solution: Ghamizi et al.~\cite{ghamizi2022adversarial} observed weighing tasks during MTL helps increase adversarial robustness. Therefore, a penalty term can be added to the merging objective~(cf. Eq.~\eqref{eq:model_merging_goal}) and tasks weighed to minimize transfer threat. Therefore, by trading a modest share of MTL accuracy for adversarial robustness, robust MM could be achieved.
However, since higher target model test accuracy and decision boundary similarity are primary drivers of transferability~\cite{yang2021trs,tramer2017space,liu2017delving,higher_test_transferability2}, we conjecture that \textit{the core objectives of MM and low adversarial transferability may fundamentally be at odds with each other}. We leave further exploration on this open for future research.

\section{Conclusion}
\label{section:conclusion}
{In this work, we explore adversarial robustness of MM, and find that the tested MM methods do not provide robustness against transfer attacks.} Instead, we show that when stronger MM methods are employed, it increases a model’s risk to transfer attacks. We statistically validate our empirical results, give theoretical explanations of the findings, and finally, discuss a potential solution to this problem. Our findings challenge the prevailing notion in the literature that MM confers adversarial robustness, and instead highlights a critical, previously overlooked vulnerability. By identifying this risk, we aim to advance the development of safer ML pipelines by understanding failure modes. We hope our work sparks future research in this important direction. 



\section*{Acknowledgment}
The work of Ankit Gangwal is supported under the ``MSCA Seal of Excellence @UNIPD'' program with the contribution of ``Fondazione Cassa di Risparmio di Padova e Rovigo'' for his ``COMPASS'' project.

\section*{Ethical Considerations}
 Given the potential implications of our finding for deployed ML systems and user-facing applications, we have taken multiple steps to ensure the ethical handling and responsible dissemination of our results. All experiments were conducted in controlled environments using publicly available datasets and models. At no point did we target production systems, external APIs, or third-party applications. No sensitive, proprietary, or human-related data were involved in this study. We have adhered to standard ethical research guidelines to ensure that the dissemination of this work minimizes harm and promotes positive impact through increased model robustness and awareness.


\bibliographystyle{splncs04}
\bibliography{bib}

\appendix
\setcounter{table}{0}
\counterwithin{table}{section}
\renewcommand{\thesection}{\Alph{section}}%
\setcounter{figure}{0}
\counterwithin{figure}{section}
\renewcommand{\thesection}{\Alph{section}}%
\setcounter{equation}{0}
\renewcommand{\theequation}{\Alph{section}.\arabic{equation}}

\section{Summary of Notation and Organization of Appendix}
\begin{table}[!htbp]
	\begin{center}
		\caption{A summary of the notations used in our work.}
		\label{tab:notations}
		\resizebox{\linewidth}{!}{%
			\begin{tabular}{clcl}
				\toprule
				\textbf{Notation} & \multicolumn{1}{c}{\textbf{Description}} & \textbf{Notation} & \textbf{Description} \\
				\midrule
				$t$ & $t^{\text{th}}$ fine-tuning task & $t_\gamma$ & Number of iterations for $\gamma$ \\
				$\mathcal{D}^t_{\mathit{tr}}$ & Training dataset & $t_s$ & Test statistic for $t$-test \\
				$\mathcal{D}^t_{\mathit{te}}$ & Test dataset & $W$ & Test statistic for Wilcoxon test \\
				$T$ & Total number of tasks & $d$ & Cohen's $d$ \\
				$m$ & MM method & $r$ & Wilcoxon's $r$ \\
				$\mathcal{M}$ & Space of MM methods & $\alpha_\gamma$ & Step size for $\gamma$ \\
				$\mathbf{x}$ & Clean input  & $\epsilon$ & Perturbation budget \\
				$\tilde{\mathbf{x}}$ & Perturbed input & $\varnothing$ & Null set \\
				$\gamma$ & Attack method & $\Delta$ & ASR difference \\
				$\Gamma$ & Space of attack methods & $\boldsymbol{\Delta}$ & ASR difference tensor \\
				$\mathbf{A}_\gamma^t$ & Transfer ASR matrix for $\gamma,\, t$ & $\mathring{\boldsymbol{\Delta}}$ & ASR difference set \\
				$\mathbf{R}_\gamma^t$ & Mean transfer rate for $\gamma,\, t$ &$\overline{R}_{\gamma}^{t,\,s}$ & Mean transfer rate for surrogate $s$ \\
				$\mu$ & Mean & $[n]$ & $\{1,\dots, n\}$ \\
				$\mu_{1/2}$ & Median & $\alpha$ & Significance level \\
				\bottomrule
		\end{tabular}}
	\end{center}
\end{table}
\appendixname~\ref{section:statistical_process_figure}  elucidates and illustrates the statistical testing procedure employed to validate the hypotheses proposed. 
 \appendixname~\ref{section:experimentalsetting_appendix} provides the experiment and hyperparameter details omitted from the main text. \appendixname~\ref{section:proof} details the proof alluded in \S\ref{subsection:hyp3_discussion}. {\appendixname~\ref{section:ensemble_discussion} provides an alternate explanation for $\mathcal{H}_3$.} {Full per-attack statistical tests, complete TSR/ASR matrices, targeted attack results, cross-architecture surrogate evaluation, independently fine-tuned surrogate results, and query-based attack details are provided in the separate Supplementary Appendix document}\footnote{{\hyperref{https://tinyurl.com/mnrl-supplementary}{}{}{https://tinyurl.com/mnrl-supplementary}}}.

\section{Statistical Testing Procedure}
\label{section:statistical_process_figure}
The three step-statistical testing procedure we used is elucidated below:
\begin{enumerate}
    \item \textbf{Normality Check.} We first check for normality of the sample, i.e., ASR differences, using Shapiro-Wilk test~\cite{shapiro1965analysis}. Here, the null hypothesis is that the sample is normal, while the alternative is the negation of the null hypothesis.
    \item \textbf{Statistical Test.} A paired $t$-test is employed in case the sample is normal, and Wilcoxon's signed rank test otherwise. Specifically,
    \begin{enumerate}
        \item if the null hypothesis of normality is not rejected, i.e., $p \geq \alpha$ (where $\alpha$ is the significance level), we perform a one-sample $t$-test~\cite{student1908probable} under the null hypothesis: $\mu \leq 0$ versus the one-sided alternative: $\mu > 0$ with significance level $\alpha$. 
        \item if the normality assumption is rejected ($p < \alpha$), we instead use the Wilcoxon signed-rank test~\cite{wilcoxon1992individual}, with the null hypothesis: $\mu_{1/2} \leq 0$ against the one-sided alternative: $\mu_{1/2} > 0$, using the same significance level $\alpha$, where $\mu_{1/2}$ denotes the median.
    \end{enumerate}
    
    \item \textbf{Benjamini–Hochberg (BH) correction~\cite{benjamini1995controllingfdr}.} To correct for multiple comparisons, we apply the BH correction procedure to the obtained $p$-values, which controls the False Discovery Rate (FDR) $Q_e$. Here, $Q_e = \mathbb{E}(Q)$, where $Q$ is the proportion of false discoveries among discoveries.
\end{enumerate}

For all tests, we set $\alpha = Q_e=  0.05 = 5 \times10^{-2}$. Finally, to evaluate the strength of the effect in the hypotheses, we also calculate the effect size for each statistical test performed. 
\begin{wrapfigure}{r}{0.5\linewidth}
	\centering
	\vspace{-1em}
	\includegraphics[width=\linewidth]{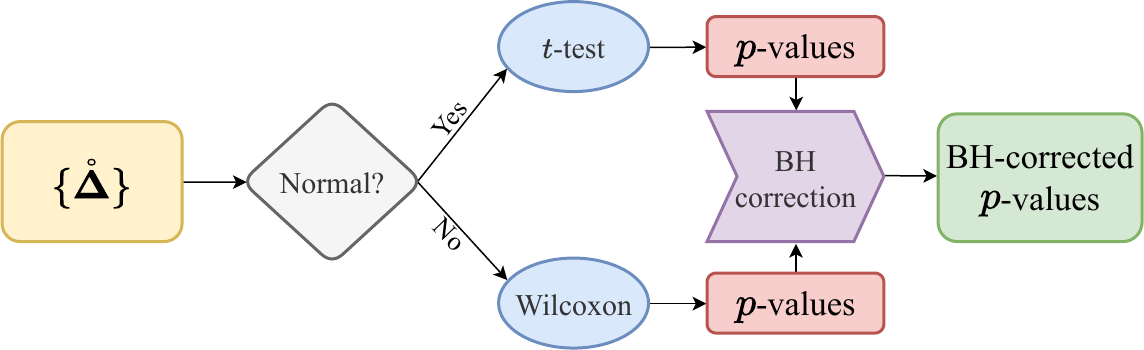}
	\caption{General statistical testing procedure for investigating transferability in MM. $\{\mathring{\boldsymbol{\Delta}}\}$  denotes set of  sets consisting of ASR differences.}
	\label{fig:general_statistical_process}
	\vspace{-1em}
\end{wrapfigure}
For paired $t$-test, we calculate the effect size $d$ using Cohen's $d$~\cite{cohen2013statistical}, i.e., $d = \frac{\bar{X}_{\mathcal{C}}}{\sigma_{\mathcal{C}}},$ where $\bar{X}_{\mathcal{C}}$ and $\sigma_{\mathcal{C}}$ are the mean and standard deviation of sample $\mathcal{C}$ consisting of  paired differences. For Wilcoxon signed rank test, we calculate the effect size $r$ as follows~\cite{tomczak2014need_for_effect_size}:
$r = \frac{Z}{\sqrt{N}},$ where $Z$ is the standardized test statistic from the Wilcoxon test and $N$ is the number of non-zero differences in the sample. \textbf{A Cohen's $d$ in the range $[0.1, 0.5)$, $[0.5, 0.8)$, and $[0.8, \infty)$ indicates a small, medium, and large effect respectively~\cite{cohen2013statistical}. Likewise, a Wilcoxon's $r$ in the range $[0.1, 0.3)$, $[0.3, 0.5)$, and $[0.5, 1]$ indicates a small, medium, and large effect respectively~\cite{tomczak2014need_for_effect_size}}. Values below $0.1$ are generally considered negligible.

\begin{notebox}{Note on Confirmatory vs. Exploratory Analysis}
	Hypothesizing After Results are Known~(HARKing) is the questionable practice of creating/changing hypothesis after seeing the patterns in the data and falsely presenting it as an a prior hypothesis~\cite{kerr1998harking}. It leads to inflated false positives and problems in reproducibility. We note that most of our hypotheses are confirmatory in nature, i.e., we hypothesized before ever looking at the data/results. However, we found WA to be an outlier while we performed our tests. Therefore, the statistical analysis for $\mathcal{H}_3$: WA being the most vulnerable MM method is more exploratory in nature than confirmatory. We also note that the statistical analysis performed in \S\ref{subsection:hyp3_discussion} for why WA is most vulnerable is confirmatory, i.e., we hypothesized before performing the statistical analysis.
\end{notebox}

\section{Experimental Details}
\label{section:experimentalsetting_appendix}

\begin{table}[H]
\centering
\caption{Model accuracies (\%) on seven tasks.}
\label{tab:performance_vitbase32}
\footnotesize
\setlength{\tabcolsep}{3pt}
\resizebox{\linewidth}{!}{%
\begin{tabular}{l|ccccccc|c}
\toprule
{Model}   &  {Cars}  &  {MNIST} &  {EuroSAT}  &  {GTSRB}  &  {SVHN}  &  {RESISC45}  &  {DTD}  & \textbf{Avg.}  \\
\midrule
{Pretrained (ViT)}  & 59.7  & 48.3 & 45.5 & 32.6 & 31.6 & 60.7 & 44.0 & 46.1 \\
{Pretrained (RN50)} & 57.0  & 61.1 & 33.6 & 36.1 & 33.8 & 54.1 & 43.2 & 45.6 \\
{Individual (ViT)}  & 77.7  & 99.7 & 99.9 & 98.7 & 97.5 & 96.0 & 79.4 & 92.7 \\
{Individual (RN50)} & 76.0  & 99.6 & 97.6 & 98.8 & 96.8 & 93.4 & 67.5 & 89.9 \\
{WA}       & 63.3  & 87.3 & 72.5 & 53.0 & 64.5 & 71.8 & 50.9 & 66.2 \\
{TA}       & 58.1  & 97.4 & 80.8 & 72.4 & 81.4 & 69.1 & 52.5 & 73.1 \\
{TM}       & 65.3  & 96.3 & 76.0 & 70.0 & 81.4 & 75.8 & 55.1 & 74.3 \\
{AM}       & 67.9  & 99.4 & 87.7 & 75.7 & 86.4 & 78.8 & 57.3 & 79.1 \\
\midrule
{WA+RS}   & 65.2 & 97.8 & 96.2 & 82.3 & 75.2 & 86.2 & 69.5 & 81.8 \\
{TA+RS}   & 61.1 & 98.6 & 95.6 & 86.7 & 87.2 & 83.2 & 67.7 & 82.9 \\
{TM+RS}   & 66.0 & 98.3 & 96.2 & 86.5 & 85.7 & 86.8 & 69.7 & 84.2 \\
{AM+RS}   & 68.5 & 99.5 & 96.4 & 87.2 & 89.4 & 87.5 & 71.6 & 85.7 \\
\bottomrule
\end{tabular}%
}
\end{table}

\subsection{Datasets}
\label{subsection:dataset}
Following prior work on TA~\cite{task_arithmetic}, TM~\cite{ties_merging}, AM~\cite{ada_merging} and RS~\cite{representation_surgery},  we conducted our experiments on the following seven benchmark datasets:
\begin{itemize}
    \item \textbf{Cars~\cite{cars}.} The Stanford Cars dataset consists of 16,185 images from 196 car classes, split equally between training and test sets.
    \item \textbf{RESISC45~\cite{resisc}.} A remote sensing scene classification benchmark with 45 classes, each containing 700 images (256$\times$256 resolution), totaling 31,500 images.
    \item \textbf{EuroSAT~\cite{eurosat}.} A geospatial land use dataset containing 27,000 Sentinel-2 satellite images annotated into 10 classes.
    \item \textbf{SVHN~\cite{svhn}.} The Street View House Numbers dataset comprises over 600,000 real-world color images of house numbers, divided into 10 classes.
    \item \textbf{GTSRB~\cite{gtsrb}.} The German Traffic Sign Recognition Benchmark features more than 50,000 images in 43 classes, exhibiting diverse lighting and background conditions.
    \item \textbf{MNIST~\cite{mnist}.} A widely used dataset of 70,000 grayscale images of handwritten digits in 10 classes, with 60,000 for training and 10,000 for testing (28$\times$28 resolution).
    \item \textbf{DTD~\cite{dtd}.} The Describable Textures Dataset contains 5,640 labeled texture images across 47 categories, with approximately 120 images per class (ranging from 300$\times$300 to 640$\times$640 pixels).
\end{itemize}

\subsection{Hyperparameters}
\label{subsection:hyperparameters}
We follow prior works~\cite{ti_fgsm,ni_fgsm} and set the perturbation budget $\epsilon$, step size $\alpha_\gamma$ and number of iteration $t_\gamma$ to $16/255$, $1.6/255$ and $10$ to generate adversarial examples. For NI-FGSM, we use the default decay factor $\mu_d=1$.  However, we observed degraded performance in case of MNIST with this configuration, since it is a single-channel dataset, as opposed to other datasets, which are three-channel. Therefore, we follow Madry et al.~\cite{madry2018towards} and choose $\epsilon=0.3$, $\alpha_\gamma=0.03$ and $t_\gamma=10$ for MNIST. In line with the original implementation of TI-FGSM, we choose the Gaussian kernel with kernel size $k= 15$ and standard deviation $\sigma=\frac{k}{\sqrt{3}}$. For Square attack, we fix query budget to 500 and size of square $p=0.8$. Finally, for targeted attacks, we randomly select the label to perform targeted attacks. We follow experimental setup of RS~\cite{representation_surgery} to obtain RS-augmented merged models, and run all experiments for 500 iterations with a batch size of 16. The rank for the adapter module is set to 16 by default. We obtain the pretrained ResNet-50 model from \texttt{open\_clip} library, and fine-tune the model for 10 epochs. We set the base learning rate to $10^{-3}$, encoder learning rate to $10^{-4}$, with the weight decay $0.1$. Finally, we use AdamW as the optimizer, clip gradients to a max norm of 1.0, and use cross-entropy as the loss.

\section{Proof}
\label{section:proof}
\begin{lemma}[Direction that maximizes average cosine similarity]
Let $v_1,\ldots,v_T\in\mathbb{R}^d$ be nonzero vectors and define the unit directions
$w_t \coloneqq v_t/\|v_t\|$ and their mean $\bar w \coloneqq \tfrac{1}{T}\sum_{t=1}^T w_t$.
For any nonzero $x\in\mathbb{R}^d$, the average cosine similarity
\[
F(x)\;\coloneqq\;\frac{1}{T}\sum_{t=1}^T \cos(x,v_t)
\;=\;\frac{1}{T}\sum_{t=1}^T \frac{\langle x,v_t\rangle}{\|x\|\,\|v_t\|}
\]
satisfies
\[
F(x)\;=\;\Big\langle \frac{x}{\|x\|},\,\bar w\Big\rangle
\;\le\;\|\bar w\|
\]
with equality if and only if $x$ is a positive scalar multiple of $\bar w$ whenever $\bar w\neq 0$.
Consequently, every maximizer has direction of $\bar w$, and the maximum value of $F$ equals $\|\bar w\|$.
If $\bar w=0$, then $F(x)=0$ for all $x\neq 0$.
\end{lemma}

\begin{proof}
Since $v_t\neq 0$, we can write $v_t=\|v_t\|\,w_t$ with $\|w_t\|=1$. Then
\[
F(x)\;=\;\frac{1}{T}\sum_{t=1}^T\frac{\langle x,\|v_t\|\,w_t\rangle}{\|x\|\,\|v_t\|}
\;=\;\Big\langle \frac{x}{\|x\|},\,\frac{1}{T}\sum_{t=1}^T w_t\Big\rangle
\;=\;\Big\langle \frac{x}{\|x\|},\,\bar w\Big\rangle.
\]
By Cauchy–Schwarz,
$\langle x/\|x\|,\bar w\rangle \le \|x/\|x\|\|\,\|\bar w\|=\|\bar w\|$,
with equality \textit{iff} $x/\|x\|$ is parallel to $\bar w$ and with nonnegative alignment at the maximizer. This yields the claim above. If $\bar w=0$, then $F(x)=0$ for all $x\neq 0$.
\end{proof}

\begin{corollary}[Unit-vector case]
If each $v_t$ is unit, then $w_t=v_t$ and $\bar w=\tfrac{1}{T}\sum_{t=1}^T v_t$. The unit maximizer is $x^\star=\bar w/\|\bar w\|$ when $\bar w\neq 0$, and the maximum equals $\|\bar w\|$.
\end{corollary}

\begin{corollary}[Application to Eq.\ \eqref{eq:wa_conjecture}]
Let $g_t(\mathbf{x})\coloneqq\nabla_{\mathbf{x}}\mathcal{L}(f_{\boldsymbol{\theta}_t}(\mathbf{x}),\mathbf{y})$ and suppose
\[
\nabla_{\mathbf{x}}\mathcal{L}\bigl(f_{\boldsymbol{\theta}_{mtl}^{\text{WA}}}(\mathbf{x}),\mathbf{y}\bigr)
\;\approx\;\frac{1}{T}\sum_{t=1}^T \frac{g_t(\mathbf{x})}{\|g_t(\mathbf{x})\|},
\]
as stated in Eq.\ \eqref{eq:wa_conjecture}. Then by the lemma, the WA gradient direction approximately maximizes the average cosine similarity with the individual gradients $\{g_t(\mathbf{x})\}_{t=1}^T$, and the maximal average cosine is approximately the norm of the average of the normalized gradients.
\end{corollary}

\section{Alternate Perspective on Explanation for \texorpdfstring{$\mathcal{H}_3$}{H3}}
\label{section:ensemble_discussion}
{Eq.~\eqref{eq:ctl} shows how WA can approximate logit ensemble of individual models. This is informative, since there exists a line of research which uses ensemble of models to defend against adversarial examples~\cite{kariyappa2019improving}. From this, one would expect WA to be more robust to transfer attacks, as they approximate ensembles. However, the key constraint for an ensemble defense to work is that the constituent models should have misaligned or orthogonal gradients, i.e., the models should be diverse~\cite{kariyappa2019improving}. If not, an adversarial example fooling one constituent model could also fool another constituent model, leading to failure of ensemble defense. We can think of WA as an operation that collapses constituent models (in ensemble setting) to a single model. As the operation is averaging, the collapsed model maintains salient features and gradient information from each constituent models, and leads to maximization of non-diversity between the collapsed model and other constituent models, making the collapsed ensemble, i.e., WA model maximally vulnerable to attacks from other constituent models. This gives an alternate perspective on explanation for $\mathcal{H}_3$ and why WA is found to be the most vulnerable to transfer attacks, despite being the weakest MM method in terms of accuracy.}

\end{document}